\documentclass{acm_proc_article-sp}

\usepackage{times}
\usepackage{subfigure}
\usepackage{paralist}
\usepackage[numbers]{natbib}
\usepackage{algorithm}
\usepackage{algorithmic}
\usepackage{amsmath,amssymb,xspace,Tabbing,bbm,array}
\usepackage{color,mathtools}
\usepackage{graphicx}
\usepackage[hyphens]{url}
\usepackage{listings}
\usepackage{multirow}
\usepackage{listings}
\usepackage{pgfplots}\pgfplotsset{compat=1.4}

\newcommand{\Cpp}{C\kern-0.05em\texttt{+\kern-0.03em+}}

\newcommand{\ConceptCpp}{ConceptC\kern-0.05em\texttt{+\kern-0.03em+}}

\lstdefinestyle{basic}{showstringspaces=false,
                       columns=fullflexible,
                       language=C++,
                       escapechar=@,xleftmargin=1pc,%
                       basicstyle=\small\sffamily,
                       commentstyle=\mdseries,
                       moredelim=**[is][\color{white}]{~}{~},
                       morekeywords={concept,
                                     unsigned,
                                     concept_map,
                                     requires,
                                     axiom,
                                     late_check},
literate={->}{{$\rightarrow\;$}}1 {<-}{{$\leftarrow\;$}}1 {=>}{{$\Rightarrow\;$}
}1,
}
\newcommand{\B}[1]{{\bf #1}}
\newcommand\beginpf{\noindent \emph{Proof: }}
\newcommand\eprf{\hfill$\Box$}

\newcommand\cS{\mathcal{S}}

\newtheorem{ass}{Assumptions}

\newtheorem{prop}{Proposition}

\begin{document}

\title{Sparse Quantile Huber Regression for Efficient and Robust Estimation}

\numberofauthors{4} 
%
\author{
%
%
\alignauthor
Aleksandr Aravkin\\
       \email{saravkin@us.ibm.com}
\alignauthor
Anju Kambadur\\
       \email{pkambadu@us.ibm.com}
\alignauthor 
Aur\'{e}lie C. Lozano\\
       \email{aclozano@us.ibm.com}
\and  
\alignauthor 
Ronny Luss\\
    \email{rluss@us.ibm.com}\\\vspace{.2 in}
       \affaddr{IBM T.J. Watson Research Center}\\
       \affaddr{1101 Kitchawan Rd.}\\
       \affaddr{Yorktown Heights, NY 10598}
   }
\additionalauthors{}
\date{30 July 1999}

\maketitle

\begin{abstract}

We consider new formulations and methods for sparse quantile regression in the high-dimensional setting. 
Quantile regression plays an important role in many applications, including
outlier-robust exploratory analysis in gene selection. 
In addition, the sparsity consideration in quantile regression enables the exploration of the entire conditional distribution of the response variable given the predictors and  therefore yields a more comprehensive view of the important predictors.
We propose a generalized OMP algorithm for variable selection, taking the misfit loss to be either the traditional quantile loss or a smooth version we call {\it quantile Huber}, and compare the resulting greedy approaches with convex sparsity-regularized 
formulations. 
We apply a recently proposed interior point methodology to efficiently
solve all convex formulations as well as convex subproblems in the generalized
OMP setting, provide theoretical guarantees of consistent estimation, and 
demonstrate the performance of our approach using empirical studies of simulated and
genomic datasets.

\end{abstract}

\section{Introduction}
\label{sec:introduction}
%
%
Traditionally, regression analyses focus on establishing the relationship
between the explanatory variables and the conditional \textit{mean} value of
the response variable.
In particular, these analyses use the $\ell_2$-norm (least squares) of the residual
as the loss function, together with optional regularization functions.
Least squares-based methods are sufficient when the data is homogenous;
however, when the data is heterogeneous, merely estimating the conditional mean
is insufficient, as estimates of the standard errors are often biased.
To comprehensively analyze such heterogeneous datasets, \textit{quantile
regression}~\cite{KB78} has become a popular alternative to least squares-based
methods.
In quantile regression, one studies the effect of explanatory variables on the
entire conditional distribution of the response variable, rather than just on its mean value. 
Quantile regression is therefore well-suited
to handle heterogeneous datasets~\cite{Buchinsky:1994}.
A sample of recent works in areas such as computational biology \cite{Zou08},
survival analysis~\cite{KG01}, and economics~\cite{KH01} serve as testament to
the increasing popularity of quantile regression.
Furthermore, quantile regression is \emph{robust to
outliers}~\cite{Koenker:2005}: the quantile regression loss is a piecewise
linear ``check function'' that generalizes the absolute deviation loss for
median estimation to other quantiles, and shares the robustness properties of
median estimation in the presence of noise and outliers.
Finally, quantile regression has excellent computational
properties~\cite{Koenker:1997}.

We study the application of quantile regression for
\emph{high-dimensional sparse models}, where the number of variables $p$ far
exceeds the sample size $n$ ($p\gg{}n$) but the number of significant
predictors $s$ for each conditional quantile of interest is assumed to be much
smaller than $p$ ($p\gg{}s$).
Sparsity plays an important role in many applications. 
For example, in the field of compressive sensing, sparsity promoting programs
allow exact recovery of sparse and compressible under-sampled signals under
certain assumptions on the linear
operator~\citet{CandesTao2005,Donoho2006_CS}.  
In many problem domains --- natural image processing, seismic imaging,
and video --- sparse regularization improves
recovery~\cite{neelamani2010esf,MCA3-2005,Fornasier:06,wakin2006civ}.  
In statistics, sparsity is often used to find the most parsimonious model that
explains the
data~\cite{Zou2006,VariableSelection_JMLR_2003,AR_LASSO_2007,Lasso1996}.  

The most popular technique used to enforce sparsity is that of regularization
with a sparsity-inducing norm; for example, the $\ell_1$
penalty \cite{Lasso1996} on the coefficient vector is often used.
Such regularization has been applied to various loss functions
other than $\ell_2$ loss, including logistic regression~\cite{Ng:2004} and
$\ell_1$-norm support vector machines~\cite{Zhu:2004}, among others. Algorithms
for learning these sparse models include gradient methods~\cite{Beck2009} and
path following methods~\cite{Ross2007}.
More recently, greedy selection methods such as Orthogonal Matching
Pursuit (OMP)~\cite{Mallat:1993} have received considerable
attention~\cite{Zhang:2008, Jalali12} as an alternative to the
$\ell_1$-penalized methods. 
In this paper, we show that these methods are also directly applicable to
other important regression problems, and demonstrate improved performance
of these formulations in the recovery of sparse coefficient vectors. 
%
%
%

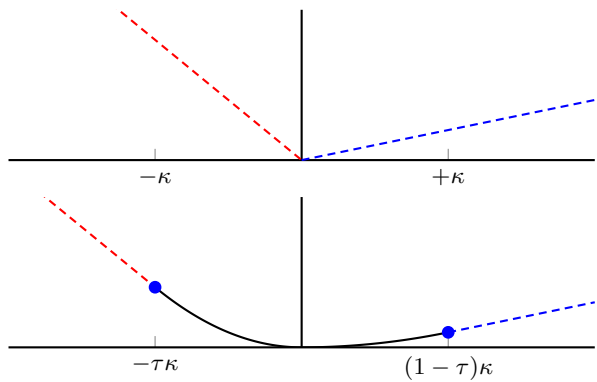
\begin{figure}
\begin{tikzpicture}
  \begin{axis}[
    thick,
    width=.44\textwidth, height=2cm,
    xmin=-2,xmax=2,ymin=0,ymax=1,
    no markers,
    samples=100,
    axis lines*=left, 
    axis lines*=middle, 
    scale only axis,
    xtick={-1,1},
    xticklabels={$-\kappa$,$+\kappa$},
    ytick={0},
    ] 
  \addplot[color = red, domain=-2:0, densely dashed]{-0.8*x};
  \addplot[color = blue, domain=0:2, densely dashed]{0.2*x};
  \end{axis}
\end{tikzpicture}
\begin{tikzpicture}
\vspace{-.05 in}
  \begin{axis}[
    thick,
    width=.44\textwidth, height=2cm,
    xmin=-2,xmax=2,ymin=0,ymax=1,
    no markers,
    samples=100,
    axis lines*=left, 
    axis lines*=middle, 
    scale only axis,
    xtick={-1,1},
    xticklabels={$-\tau\kappa$,$(1-\tau)\kappa$},
    ytick={0},
    ] 
\addplot[red,domain=-2:-1,densely dashed]{-0.8*x-0.8*.5};
\addplot[domain=-1:0]{.5*.8*x^2};
\addplot[domain=0:1]{.5*.2*x^2};
\addplot[blue,domain=+1:+2,densely dashed]{0.2*x-0.2*.5};
\addplot[blue,mark=*,only marks] coordinates {(-1,.8*.5) (1,.2*.5)};
  \end{axis}
\end{tikzpicture}
\caption{Top: quantile penalty. Bottom: quantile Huber penalty.\label{PenaltyPics}
}
\end{figure}

%
Recently, quantile regression with an $\ell_1$
penalty~\cite{Belloni:2011,Ross2007,Li2008} and with non-convex penalties (smoothly clipped
absolute deviation and minimax concave)~\cite{Wang:2012} was studied
in a high dimensional setting.
Our interest in quantile regression is two-fold: (1) to find new
efficient and scalable algorithms to quickly compute these robust sparse
models and (2) to apply it to high-dimensional biological datasets, where 
such modeling is highly relevant.
%
Therefore, we extend~\citet{Belloni:2011} to consider \textit{greedy} and
\textit{convex} formulations for sparse quantile regression in a
high-dimensional setting.
Our main contributions are:
\begin{list}{\labelitemi}{\leftmargin=1em}
\item We generalize the classic quantile check function to a {\it quantile
Huber} penalty (see Figure~\ref{PenaltyPics}). 
In many cases, this formulation holds a significant advantage since the
classic quantile check function attempts to fit a portion of the data
exactly. 
Such exact fitting, while useful in some noiseless settings, is undesirable 
in the presence of noise and outliers. While some smoothing of 
the quantile loss has been proposed in the past for computational efficiency~\cite{Zheng2011},
this is not our motivation, since the computational efficiency of our approach is not 
significantly affected by smoothness. We show in our experiments that the 
quantile Huber regression penalty is able to produce better results in simulated experiments.
\item We propose a generalized OMP algorithm for sparse quantile regression 
and extend it for the quantile Huber loss function.
Using the greedy OMP algorithm instead of $\ell_1$-penalized algorithms allows
us to develop efficient, scalable implementations. 
Furthermore, the greedy OMP algorithm exhibits significant recovery improvements 
over $\ell_1$-penalized algorithms in scenarios where quantiles other than 
50\% must be considered to capture all relevant predictors.
\end{list}

%
%
To demonstrate the significance of our contributions, we compare and contrast
four formulations:
\begin{enumerate}
\item quantile check loss with $\ell_1$ penalty ($\ell_1$-QR)
\item quantile check loss with $\ell_0$ constraint ($\ell_0$-QR)
\item quantile Huber loss with $\ell_1$ penalty ($\ell_1$-QHR) 
\item quantile Huber loss with $\ell_0$ constraint ($\ell_0$-QHR). 
\end{enumerate}

In particular, we present methods for convex problems of the form 
\[
\min \sum_{i=1}^n \rho(b_i-A_i^Tx) + \lambda \|x\|_1
\]
where $\rho$ can be the quantile or quantile Huber penalty (approaches 1 and 3), 
as well as a generalized OMP algorithm for nonconvex problems of the form
\[
\min \sum_{i=1}^n \rho(b_i-A_i^Tx) \quad\mbox{subject to}\quad \|x\|_0 \leq s
\]
to address approaches 2 and 4. The same optimization approach we use
for 1 and 3 is also used to solve the convex subproblems required for 2 and 4. 
In order to optimize all convex formulations and subproblems, 
we exploit a dual representation of the 
quantile-based loss functions and use a recently proposed interior point (IP) approach. 
IP methods directly optimize the Karush-Kuhn-Tucker (KKT) optimality conditions, 
and converge quickly even when the regression problems
are ill-conditioned.  The particular IP approach we chose allows us 
to easily formulate and solve all the problems of interest 
in order to compare their performance.   
%
%
Our experiments demonstrate that, in a majority of the cases, $\ell_0$-QHR performs 
best; in cases where $\ell_1$-QHR performs comparably, we show that $\ell_0$-QHR 
has properties such as quick convergence that make it 
more suitable in high-dimensional settings.
%
%

The rest of this paper is organized as follows.
In Section~\ref{sec:formulation}, we explain the different quantile-based loss
functions and penalties that we use in the rest of the paper.
In Section~\ref{sec:OMP}, we present the generalized OMP algorithm to solve the
$\ell_0$-constrained quantile and quantile Huber loss functions.
In Section~\ref{sec:SIP}, we present the dual representation of the
quantile-based loss functions and briefly explain the generalized interior
point approach used to solve these formulations.
In Section~\ref{sec:theory}, we discuss the numerical and statistical
convergence of quantile Huber loss functions with $\ell_0$ constraints; the
convergence with $\ell_1$ penalties can be obtained by adapting the asymptotic
analysis from~\citet{Belloni:2011}.
Finally, in Section~\ref{sec:experiments}, we present extremely encouraging 
experiments on both synthetic and genomic data that show the effectiveness 
of both the quantile Huber loss function and the OMP approach.

\section{Problem Formulation}
\label{sec:formulation}
\subsection{Notation and background}
Let $A \in \mathbb{R}^{n \times p}$ denote the predictor matrix, whose rows are
$p$-dimensional feature vectors for $n$ training examples. 
$A_{ij}$ corresponds to the $j^{\rm
th}$ feature of the $i^{\rm th}$ observation.  
Let $b \in \mathbb{R}^{n}$ denote the response vector, with $b_i$ as the
$i^{\rm th}$ observation.
Quantile regression assumes that the $\tau$-th quantile is given by
\begin{equation}
F^{-1}_{b|A}(\tau) = A \bar{x}_\tau \label{eq:Reg}
\end{equation}
where $\bar{x}_\tau \in \mathbb{R}^p$ is the coefficient vector that we want 
to estimate and $F_{b|A}$ is the cumulative distribution function for a multivariate 
random variable with the same distribution as $b|A$.
Let $r=b-Ax$ be the vector of residuals. Quantile regression is traditionally solved
using the following ``check-function:'' 
\[c_{\tau}(r)= (-\tau + 1\{r\geq 0\})
r,\]
where the operations are taken element-wise; note that setting $\tau=0.5$
yields the Least Absolute Deviation (LAD) loss.

\subsection{A quantile Huber loss}
The quantile loss has two important modeling features: first, it is robust to
outliers, and second, it sparsifies the residual $b- Ax$ because of the
behavior of the loss at the origin. 
If this second behavior is not expected or desired, an asymmetrical Huber can
be designed by rounding off the function at the origin. 
This penalty, which we call {\it quantile Huber}, still maintains the asymetric
slopes required from the quantile penalty outside the interval $[-\kappa\tau,
\kappa(1-\tau)]$ (see right panel of Figure~\ref{PenaltyPics}).  
For a scalar $x$, the explicit formula is given by
\begin{equation}
\label{quantileHuber}
\small
\rho_\tau(x) = \begin{cases}
\tau|x| - \frac{\kappa \tau^2}{2} & \text{if} \quad x < -\tau\kappa\\
\frac{1}{2\kappa}x^2 & \text{if} \quad x\in [-\kappa \tau, (1-\tau)\kappa]\\ 
(1-\tau)|x| - \frac{\kappa(1-\tau)^2}{2} & \text{if}\quad x > \quad (1-\tau)\kappa
\end{cases}
\end{equation}

In this paper we assume that the true model parameters are sparse, namely that for a
given quantile $\tau$, the true model parameter $\bar x$ has a support of
cardinality $s \ll p$.
To enforce sparsity we consider both the $\ell_1$ penalized estimation
framework
\begin{equation}\label{eq:al:obj}
\hat x = \arg\min \sum_{i=1}^n \rho_\tau(b_i-A_i^Tx) + \lambda \|x\|_1
\end{equation}
and the greedy approach that \emph{approximately} solves the minimization of
$\sum_{i=1}^n \rho_\tau(b_i-A_i^Tx)$ subject to the constraint $\|x\|_0 = T,$
where $T$ is a desired level of sparsity.

\section{Generalized OMP for $\ell_0$-QHR}
\label{sec:OMP}

Orthogonal matching pursuit (OMP) is an effective method for finding sparse solutions to data fitting problems. 
Given a basis $\cS$ of size $k$, classic OMP computes the residual $r = b - A_{\cS} x_{\cS}$
(where $A_{\cS}$ is the submatrix of $A$ comprising columns in $\cS$, and $x_{\cS}$ are the corresponding model coefficients), 
and then chooses the next element to add to $\cS$ by computing 
\[
i = \arg\max_{i} |r^TA_i|\;,
\] 
that is, the index where the maximum absolute value of the projection of the residual 
is achieved. 

This approach generalizes to arbitrary loss functions $\rho$, 
including the quantile and quantile Huber loss functions. 
Generalized OMP for $\ell_0$-QHR is detailed in Algorithm~\ref{genOMP}. 

\begin{algorithm}\caption{OMP for $\ell_0$-QHR}\label{genOMP}
\noindent
{\small
{\bf Initialization:} $r=b, S^{(0)}=\emptyset$\\
{\bf For}: $k=1,\ldots$ 
{Selection Step:}\\
\begin{align}
  & r^{(k)} = b - A_{\cS^{(k)}}x^{(k)}\\
	&\label{argmax} i_{(k)}=\arg\max_i \left|\nabla \rho\left(r^{(k)}\right)^TA_i\right|\\
	&\quad \textrm{   (Maximum projection onto generalized residuals)}\nonumber\\
       & S^{(k)}=S^{(k-1)} \cup \{i_{(k)}\}\nonumber
\end{align}
{Refitting Step:}
\begin{align}
       & x^{(k)}= \arg\min_{x:x_i=0, i \notin S^{(k)}} \rho(b- Ax) \label{eq:EE} 
\end{align}
}
\end{algorithm}

When the loss $\rho$ is not differentiable, a subgradient can be used instead in~\eqref{argmax}. 
For example, for the quantile loss with parameter $\tau$, we interpret $\nabla c_\tau(r)$ as follows:
\[
\nabla c_\tau(r) := (1-\tau) r_+ + \tau r_-\;. 
\]
Note that the quantile Huber loss is differentiable, and the gradient is easily computable from~\eqref{quantileHuber}. 

The refitting step \eqref{eq:EE} can be solved using any algorithm. 
In particular, the largest submatrix $A_{\cS}$ used to solve any~\eqref{eq:EE} problem 
will be $n\times k$, with $k$ the total number of OMP steps, 
and $k$ is small when the solution is expected to be very sparse. 
This regime favors the IP method described in section~\ref{sec:SIP}, and is dominated 
by costs of forming ($O(nk^2)$) and solving ($O(k^3)$) a particular linear system discussed in section~\ref{sec:SIP}.
When used as a subroutine to generalized OMP, IP methods very rapidly, reliably, and accurately 
solve the refitting step~\eqref{eq:EE} for smooth and nonsmooth penalties $\rho$.  

\section{Efficient optimization for Q(H)R}
\label{sec:SIP}
This section describes our optimization approach to solving $\ell_1$-QR and $\ell_1$-QHR penalized problems, 
as well as the refitting step for $\ell_0$-QR and $\ell_0$-QHR. We first show how our objectives fit into a very general class of functions and give 
a description of a recently proposed IP algorithm for this class~\cite{AravkinBurkePillonetto2013}. The section concludes by detailing 
how to parallelize the required dense linear algebra operations for large-scale data using MATLAB and C++ 
 

We first note that other algorithms could be applied to these problems, but a rigorous 
comparison is out of the scope of this paper.  For example, the refitting step in OMP is smooth for quantile Huber Regression, and the 
objective of $\ell_1$-QHR fits into the well-known framework of fast gradient methods for
optimizing certain smooth plus non-smooth functions (e.g., \citet{Beck2009}).  While these methods are competitive
with what we describe, the more general framework used here efficiently solves all of our problems, 
for both smooth and nonsmooth objectives.

\subsection{Piecewise Linear Quadratic Functions}
\label{ss:PLQ}
A broad class of functions known as piecewise linear quadratic (PLQ) are of the
form  
\begin{equation}
\label{PLQrepFull}
f(x) := \sup_{u\in U} \left\langle u, Bx +b \right\rangle - \frac{1}{2} u^T M u\;,
\end{equation}
where $U := \{u : Cu \leq c\}$ is a polyhedral set, and
$M$ is positive semidefinite~\cite{RTRW}. When $0\in U$, $f(x)\ge0$ for all $x\in \mathbb{R}^p$,
and we therefore refer to these functions as {\it penalties}.   
Note that $M=0$ for any piecewise linear penalty, such as the quantile penalty
(see left panel of Figure~\ref{PenaltyPics}).  
 
The quantile penalty for $r\in \mathbb{R}^n$ can be written
\begin{equation}
\label{QuantilePenalty}
c_\tau(r) = \tau \sum_i  (r_i)_-+(1-\tau)\sum_i (r_i)_+ \;,
\end{equation}
where $x_+ = \max(0, x)$ and $x_- = \max(-x, 0)$.
The quantile penalty can be represented using notation~\eqref{PLQrepFull} by taking
\begin{equation}
\label{quantilePLQ}
M = 0, B = I,b = 0,
C = \begin{bmatrix} I \\ -I\end{bmatrix}, \text{ and } c = \begin{bmatrix} (1-\tau)\mathbf{1} \\ \tau\mathbf{1}\end{bmatrix}\;,
\end{equation}
to obtain 
\begin{equation}
\label{finalQuantile}
c_\tau(r) = \sup_{u \in [-\tau, (1-\tau)]^{n}} \left\langle u, r \right\rangle\;.
\end{equation}
By taking $M = \kappa I$, the quantile Huber penalty can be represented as
\begin{equation}
\label{finalQHuber}
\rho_\tau(r) = \sup_{u \in [-\tau, (1-\tau)]^{n}} \left\langle u, r \right\rangle- \frac{\kappa}{2}u^Tu\;. 
\end{equation}
Note that in our regression settings, $r=b-Ax$ is the residual of a linear function 
on data matrix $A$ and can be plugged into both~\eqref{finalQuantile} and~\eqref{finalQHuber} 
by taking $B=-A$ and $b$ as the response vector.


Moreover, it is easy to see that adding a sparse regularization term $\lambda \|x\|_1$ to any objective with PLQ penalty encoded by 
$C, c, B, b, M$ gives another PLQ penalty with augmented data structures $\tilde B =[B^T, I]^T$, $\tilde b=[b^T 0]^T$,
\begin{equation}
\label{SumPLQ}
\tilde C = \begin{bmatrix} C \\ I \\ -I\end{bmatrix}, 
\tilde c = \begin{bmatrix}c \\ -\lambda \mathbf{1} \\ \lambda\mathbf{1} \end{bmatrix},
\mbox{ and } \tilde M = \begin{bmatrix} M  & 0 \\ 0 & 0\end{bmatrix}\;.
\end{equation}

\subsection{Large-scale interior point approach}
\label{ss::IPapproach}
The significance of representations~\eqref{finalQuantile},~\eqref{finalQHuber}, and~\eqref{SumPLQ} is that all objectives of our
interest (including $\ell_1$-QR/QHR and the generalized OMP refitting step for $\ell_0$-QR/QHR) are
of the form~\eqref{PLQrepFull}. Many of the matrices in these PLQ representations are {\it hypersparse}; that is,
the number of non-zeros is of the order of the matrix dimension. In this section, we briefly 
review the interior point (IP) approach presented in~\citet{AravkinBurkePillonetto2013} and 
show how to exploit hypersparsity. 

The Karush-Kuhn-Tucker (KKT) optimality conditions for~\eqref{PLQrepFull} are
\begin{equation}
\label{fullKKT}
F(x, u, s, q):= 
\begin{bmatrix} B^Tu \\ b + Bx - Mu - C^Tq \\ Cu + s - c \\ Qs\end{bmatrix}
= 0\;,
\end{equation}
where $s$ is a slack variable added to the inequality constraint, and $Q=\mbox{diag}(q)$ is the dual variable 
corresponding to the resulting equality constraint. IP methods iteratively solve $F=0$
using relaxed systems $F_\mu$, obtained by 
approximating the complementarity conditions $Qs=0$ in~\eqref{fullKKT} 
by $Qs = \mu$. Specifically, IP methods solve, at each iteration, the equation defined by a first-order approximation to $F_\mu$ 
\begin{equation}
\label{NewtonIteration}
F_{\mu}^{(1)}\begin{bmatrix}\Delta x^T & \Delta u^T & \Delta s^T & \Delta q^T\end{bmatrix}^T = -F_\mu\;,
\end{equation}
where $F_{\mu}^{(1)}$ is derivative matrix of $F$ with respect $(x,u,s,q)$. Parameter $\mu$ is quickly driven to $0$ 
as the iterations proceed, so that we solve $F=0$.

IP methods are explained in many classic sources~\cite{Ye:1993,Wright:1997} and exhibit super-linear convergence.
In order to discuss efficiency, 
we give an explicit algorithm to solve~\eqref{NewtonIteration}. 
Defining 
\begin{equation}
\label{TandOmega}
T := M + C^TQS^{-1}C, \quad \Omega = B^TT^{-1}B,
\end{equation}
where $S=\mbox{diag}(s)$, the full Newton iteration~\eqref{NewtonIteration} is implemented as follows:
\begin{equation}
\label{IPiter}
\begin{aligned}
r_1 &= -s - Cu + c
&
\Delta x &= \Omega^{-1} r_4
\\
r_2 &= \mu\B{1} + Q(Cu - c)
&
\Delta u & = T^{-1}(-r_3 + B\Delta x)
\\
r_3 &= -(Bx - Mu - C^Tq)
&
\Delta q & = S^{-1}(r_2 + QC\Delta u)
\\
&\quad-b+C^TS^{-1}r_2
&
\\
r_4 &= -B^T u  +B^T T^{-1}r_3 
&
\Delta s & = r_1 - C \Delta u
\end{aligned}
\end{equation}

For $\ell_1/\ell_0$-QR/QHR using formulations~\eqref{finalQuantile}, \eqref{finalQHuber}, and~\eqref{SumPLQ}, $C$ is
simply a stack of signed identity matrices, and $Q$, $S$, and $T$ are diagonal by construction. 
The computations needed in~\eqref{IPiter} are vector additions, matrix-vector products,
matrix-matrix products, and linear solves.
%
%
Most of the operations are sparse (and fast); 
the expensive operations are 
forming $\Omega$ 
and solving for $\Delta{}x$. 
%

In the case of $\ell_0$-QR/QHR, OMP picks $k\ll{}p$ columns corresponding to a basis estimate 
for the refitting step at iteration $k$, and $B$ in~\eqref{IPiter} is simply the 
$k$-column submatrix $A_{k}\in\mathbb{R}^{n\times k}$ of data matrix $A$. 
The cost of forming $\Omega$ is  $O(nk^2)$, and the cost of
solving for $\Delta{}x$ is $O(k^3)$ at iteration $k$. 
%
%
Regarding $\ell_1$-QR/QHR, 
$B$ in~\eqref{IPiter} is augmented as shown in~\eqref{SumPLQ} and 
$\Omega = A^T T_n^{-1}A + T_p^{-1}$ ($T_n/T_p$ are appropriate submatrices of $T$). The Woodbury inversion
formula can be applied to solve for $\Delta x$
in~\eqref{IPiter}: 
\[
\Delta x = \Omega^{-1}r_4 = T_p r_4 - T_p A^T \Phi^{-1}A T_p r_4\;,
\]
where $\Phi := T_n + A T_p^{-1}A^T\in \mathbb{R}^{n\times n}$ is a square positive definite matrix
in the {\it smaller} dimension $n$ (number of samples). 
$\Phi$ requires $O(pn^2)$ operations to form and $O(n^3)$ to solve. 

The IP algorithm is thus very efficient for all formulations on problems up to a 
moderate-sized number of samples.  First order methods would typically require thousands 
of more iterations at lower complexities of $O(np)$. 
First order methods may be preferable for large sample sizes; 
however, for certain special cases, inexact IP methods have been shown to be competitive 
with state of the art first order methods for huge scale problems~\cite{Boyd2007}.
For the problems we studied, the IP framework detailed in this section was used for all 
convex formulations and convex subproblems of the generalized OMP algorithm. 
%

\subsection{Solving large-scale systems}
\label{subsubsec:large_scale}
The results in this paper (see Section~\ref{sec:experiments}) were generated
using a multi-threaded version of MATLAB, which is very efficient for basic
dense- and sparse-linear algebra operations.
However, when the datasets (and $A$) are large, MATLAB is unable
to load the datasets into its memory.
Fortunately, it is possible to isolate the operations that involve $A$
and to use MATLAB for the remaining computations~\footnote{
We assume that vectors of size $p$ and matrices of size $n\times n$
fit in memory on one machine (and hence, in MATLAB).
If this assumption does not hold, all computations must be implemented in 
distributed memory using lower-level languages such as C/C++.
}.
%
%
For large datasets, $A$ is maintained as a file handle throughout the MATLAB
code and never loaded into memory.
When operations involving $A$ are to be performed, they are dispatched to a MEX
function that contacts a parallel server to perform the operations.
Our parallel server makes use of Elemental~\cite{Poulson:2012:ENF}, a
distributed dense matrix linear algebra package that makes use of
MPI~\cite{MPI-1,MPI-2} for parallelism.
The server is capable of (optionally) caching matrices, which allows us to load
the matrix $A$ only once and minimize disk cost.
For example, when $Ax$ needs to be computed, the MEX function writes out $x$
to disk and contacts the server with a request to compute $Ax$.
The server then reads $x$ into distributed memory, computes $Ax$ (assuming $A$
was loaded earlier), and writes out the resulting vector to disk, which is read
back in by the MEX function.
Note that as $T$ is diagonal in our formulations, we can compute
$\hat{A}=AT_n^{-\frac{1}{2}}$ as a diagonal scaling operation on $A$ and use
the resulting $\hat{A}$ to form $\Phi$.

\section{Convergence and consistency of OMP for $\ell_0$-QHR} \label{sec:theory}
In this section we study the behavior of the quantile Huber OMP estimator ($\ell_0$-QHR)  in the high dimensional setting, i.e. in cases where $p\gg{}n$. 
To the best of our knowledge, matching pursuit methods have not been considered  before in the context of quantile regression, nor have they been studied theoretically.
Here we study the various components involved in securing the numerical and statistical convergence of OMP for $\ell_0$-QHR. 

\subsection{Notation} Let $L_n$ denote the empirical quantile Huber loss: \[L_n(x)=\frac{1}{n} \sum_{i=1}^n \rho_\tau(b_i-A_i^T x).\] Let $L$ denote the population loss, namely the expectation of the loss
\[L(x)=\mathbb{E}_{A,b} L_n(x).\] 
%
For any reference vector $\hat x$ we define the following three categories of indices. 
\begin{eqnarray*} 
E_0(\hat x) &=&\{ i : b_i - A_i^T \hat x < - \tau \kappa \}\\ 
E_1(\hat x) &=& \{ i : b_i - A_i^T \hat x \in [ -\kappa \tau, (1-\tau) \kappa]\}\\
 E_2(\hat x) &=& \{ i : b_i - A_i^T \hat x > (1 - \tau) \kappa \} \end{eqnarray*} 
$E_0(\hat x) \cup E_2(\hat x)$ can be seen as the set of outlying observations, while $E_1(\hat x)$ is the set of inlying observations based on $\hat x.$ Finally, let $A_E$ denote the restriction of matrix $A$ to the rows in set $E.$ 

\subsection{Numerical convergence} The numerical convergence rate of  OMP for $\ell_0$-QHR as an optimization method is guided by three properties: (i) Restricted strong convexity, (ii) Lipschitz continuity and (iii) smoothness of the quantile Huber loss. We refer
the reader to~\citet{siopt10} for the definition of these properties.
The following proposition establishes Lipschitz and smoothness properties for the quantile Huber loss. 
\begin{prop} 
The quantile Huber loss function $\rho_\tau(\cdot,b)$ is Lipschitz continous with Lipschitz
constant $\max(1-\tau,\tau).$ The quantile Huber loss function $\rho_\tau(\cdot,b)$ is smooth with smoothness constant $\frac{1}{\kappa}.$ 
\end{prop} 
\beginpf Any global bound on the derivative of the loss is a
Lipschitz constant for the loss function, and by construction of the quantile Huber loss, the maximum of the slopes of the linear portions of the loss is such a bound, namely $\max(1-\tau,\tau).$ The smoothness
constant is the second derivative of the quadratic portion of the quantile Huber loss, namely $\frac{1}{\kappa}.$
\eprf

The following proposition characterizes the numerical convergence of  OMP for $\ell_0$-QHR
\begin{prop} \label{prop:nc}
Let $\tilde x$ denote the population minimizer of the quantile huber loss. 
Assume that  the OMP  algorithm for $\ell_0$-QH is run for $k$ iterations, and produces the iterate $x^{(k)}$.  Then for any
$\epsilon>0$ such that \[k \geq \frac{2 \|\tilde x\|_1^2}{\kappa \epsilon},\] there holds \[L_n(x^{(k)})-L_n(\tilde x) \leq \epsilon.\] 
\end{prop} 
\beginpf 
Noting that the quantile Huber loss is smooth with
constant $\frac{1}{\tau}$ the result follows from theorem 2.7 in~\citet{siopt10}. 
\eprf

Exponentially better numerical convergence can be secured under {\it restricted strong convexity} (see~\citet{siopt10}[Definition 1.3]). To guarantee the latter property we need the following assumption, which is
commonly made on the entire matrix $A$ in the study of high dimensional least squares regression.
\begin{ass} \label{ass:se} (Sparse Eigenvalue on $A_{E_1 (\tilde x)}$ ). Given any positive integer $k,$ for all $\| {x-x'} \|_0 \leq k$ we require that the matrix $A_{E_1 (\tilde x)}$ satistisfies the
restricted eigenvalue property. Namely there exist $\gamma(k)>0$ such that $\frac{1}{n} { \| A_{E_1(\tilde x)} (x-x') \|}_2^2 \geq \gamma(k) \|(x-x')\|^2_2.$
\end{ass}
\begin{prop} \label{prop:ncfast} Under Assumption~\ref{ass:se}, the quantile Huber loss enjoys the restricted strong convexty property with constant $\gamma(k)$ on the set $\{x \in B (\tilde x) : \|x\|_0 \leq k + \|\tilde x \|_0\},$ from some
ball $B(\tilde x)$ centered around $\tilde x.$ 
\end{prop}
\beginpf In a small neighborhood $B(\tilde x)$ of $\tilde x$ we have \[ \sum_{i=1}^n \rho_{0.5} (b_i - A_i^T x) \geq \sum_{i \in E_1(\tilde x)} \frac{1}{2\kappa} (b_i - A_i^T x)^2. 
 \] Due to the restriced eigenvalue property of $A_{E_1(\tilde x)}$ we obtain the desired result. 
\eprf

Improved convergence rates can then be obtained as soon as  OMP for $\ell_0$-QHR reaches the region of restricted strong convexity around $\tilde x.$ 
\begin{prop} Assume that the OMP algorithm for $\ell_0$-QHR reaches the restricted strong
convexity region after $k_0$ iterations and produces the iterate $x^{(k_0)}$. Then for any $\epsilon>0$ such that \[k \geq \frac{\|\tilde x\|_0}{\kappa \gamma(k_0+k)} \log \frac{L_n(x^{(k_0)})-L_n(\tilde
x)}{\epsilon},\] there holds \[L_n(x^{(k+k_0)})-L_n(\tilde x) \leq \epsilon.\] \end{prop} 
\beginpf The proof follows by adapting the reasoning of Theorem 2.8 in~\cite{siopt10}. Specifically, careful inspection of
the proof of Theorem 2.8 in ~\citet{siopt10} reveals that restricted convexity need not hold everywhere but only in a certain neighborhood around $\tilde x.$ Thus as long as  OMP for $\ell_0$-QHR reaches the
restricted strong convexity region around $\tilde x$ and Assumption~\ref{ass:se} holds the convergence rate improves exponentially. \eprf

We note that as an extreme case, if $\kappa$ is large enough so that $\mathbf{0}$ belongs to the ball $B(\tilde x)$, then the situation reduces to the simple quadratic case and the algorithm enjoys fast convergence throughout.

\subsection{Statistical consistency}  We now briefly discuss the statistical consistency of OMP for $\ell_0$-QHR. 
A formal technical analysis is beyond the scope of this paper and will be presented in future work.
Recall that $\tilde{x}$ denote a population minimizer of the quantile Huber loss and $x^{(k)}$ the iterate output by  OMP for $\ell_0$-QHR after the $k$-th iteration. We have the following decomposition: 
\[ \begin{array}{ll}
\mathbb{E} L(x^{(k)})-L(\tilde x)&\leq \mathbb{E}|  L_n(x^{(k)})-L(x^{(k)})|+ \mathbb{E} |L_n(\tilde x)-L(\tilde x)| \\
&\quad\quad+ \mathbb{E}|L_n(x^{(k)})-L_n(\tilde x)|. 
\end{array}\]
On the right-hand side of the inequality, the convergence of the second term is guaranteed using the traditional central limit theorem, since $\tilde x $ is fixed. The third term characterizes the numerical
convergence of  OMP for $\ell_0$-QHR and was dealt with in Propositions~\ref{prop:nc} and~\ref{prop:ncfast}. The first term can be bounded using standard results from empirical process theory (e.g. \citet{bm-em-05}). For instance we get that uniformly for all $x: \|x\|_0 \leq k, \|x\|_2 \leq R,$ there holds $ \mathbb{E} L(x)-L_n(x) \leq 2 \max(1-\tau,\tau) \frac{C}{\sqrt{n}}.$  Combining all the pieces allows one to conclude that the expected quantile Huber loss of the estimate produced by OMP for $\ell_0$-QHR converges to the infimum population loss.

To guarantee the consistency in terms of the original quantile loss rather than the Quantile Huber loss, it now remains to address how well a population minimizer of the quantile Huber loss approximates a population minimizer of the original quantile loss. The early work of~\citet{clark85} sheds partial light on the relationship between both estimators (see in particular Theorem 6 in~\citet{clark85}). Subsequently, the question was fully addressed by \citet{LiW98} for the case whe $\tau=0.5$. Specifically they showed that the solution set of the Huber estimator problem is Lipschitz continuous with respect to the parameter $\kappa$, and thus that the set of the traditional $\ell_1$ estimators is the limit of the set of the Huber estimators as $\kappa \to 0.$ This result can naturally be extended to general quantile Huber and quantile regression.

\section{Experiments}
\label{sec:experiments}
In this section, we present experiments with simulated and real data, with very promising results. 
 The first subsection demonstrates the use of quantile versus quantile Huber loss in feature selection tasks.  
In the second section we study an eQTL problem to discover variations in the genome 
that are associated with the \emph{APOE} gene, a key gene for Alzheimer's disease.

\subsection{Simulations}

We employ a simulation setting similar to~\cite{Wang:2012}. The data matrix $A$ is generated in two steps. 
In the first step, an auxiliary $n \times p $ matrix $Z$ is generated from a multivariate normal distribution $N(0,\Sigma),$ 
where $\Sigma_{jk}=0.5^{|j-k|}.$ 
In the second step, we set $A_1=\Phi(Z_1),$ where $\Phi$ is the normal cumulative distribution function, 
and $A_{2:p}=Z_{2:p}.$ 
The response vector $b$ is generated according to the model
\[b =A_6 + A_{12} + A_{15} + A_{20} +0.7 A_1 \epsilon + \eta,\] where $\epsilon \sim N (0, 1)$ and $\eta \sim N(0,1),$ 
are independent of one another and of the matrix $A$. 
It is important to note that  $A_1$ impacts the conditional distribution of $b$ given the predictors,
but does not directly influence the center (mean or median) of the conditional distribution. 

We consider $p\in \{400,800\}$ and $n=300.$ The comparison methods are 
$\ell_0$-QHR (OMP with quantile Huber Loss), 
$\ell_0$-QR (OMP with the traditional check function), 
$\ell_1$-QR (quantile loss with $\ell_1$ regularization), 
$\ell_1$-QHR (quantile Huber loss with $\ell_1$ regularization), 
and Lasso (the traditional Lasso estimator with $\ell_2$ loss). We run 100 simulation runs (considering 100 different datasets). 
For each simulation run, the parameters are selected via holdout validation, using a holdout dataset of size $10n.$ 

As a measure of variable selection accuracy, we report the $F_1$ score which is the harmonic mean between precision and recall. 
Specifically, the $F_1$ score is 
\[
F_1 = \frac{2\cdot \text{Prec}\cdot \text{Rec}}{\text{Prec}+\text{Rec}}, \quad \text{Prec}=\frac{{tp}}{{tp}+{fp}}, \quad \text{Rec}=\frac{{tp}}{{tp}+{fn}},
\] 
with $tp, fp, fn$ denoting true positives, false positives, and false negatives. 

\begin{figure}[t!]
\centering
\begin{tabular}{c}
\includegraphics[width=0.35\textwidth]{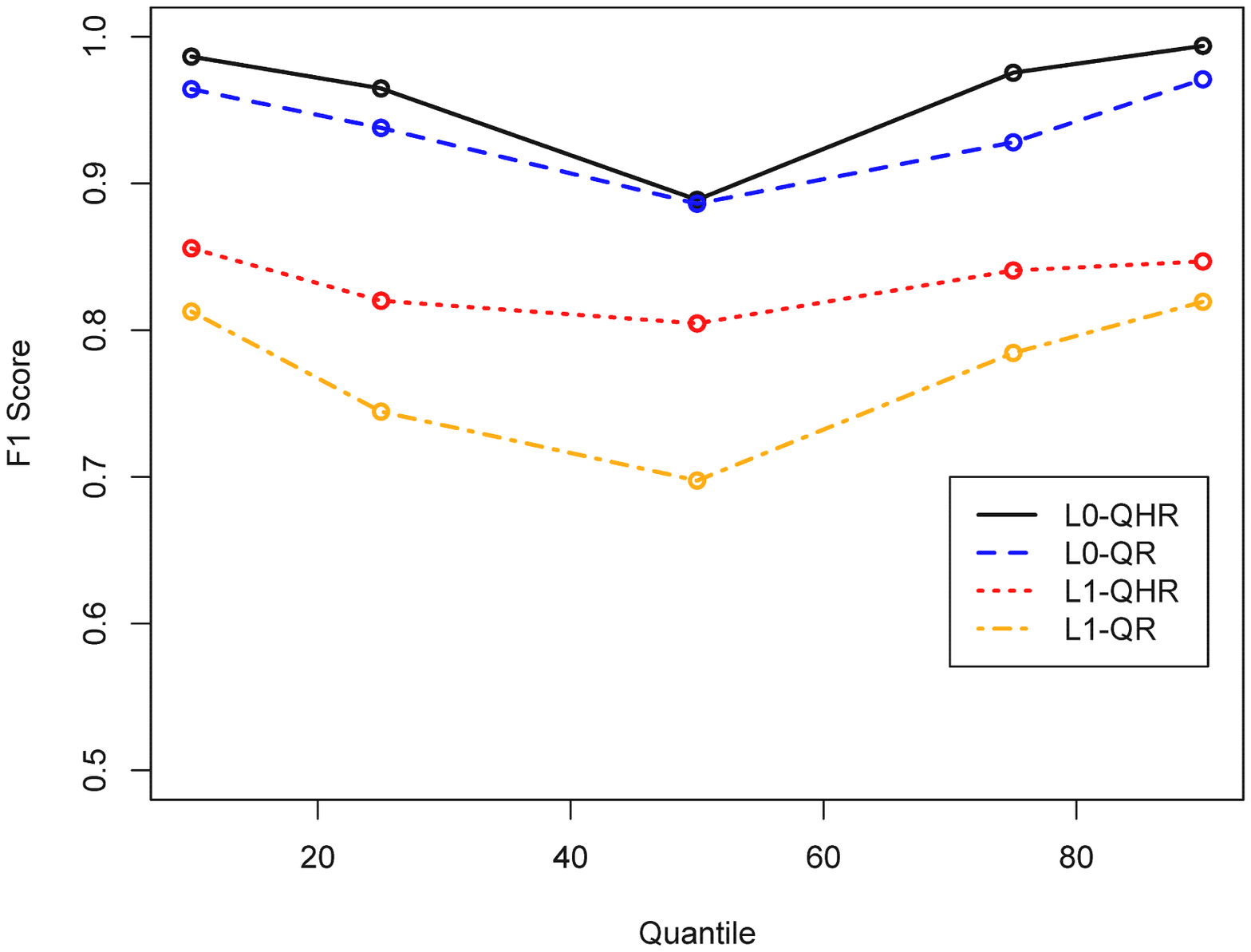} \\
$n=300, p=400$ \\
 {\includegraphics[width=0.35\textwidth]{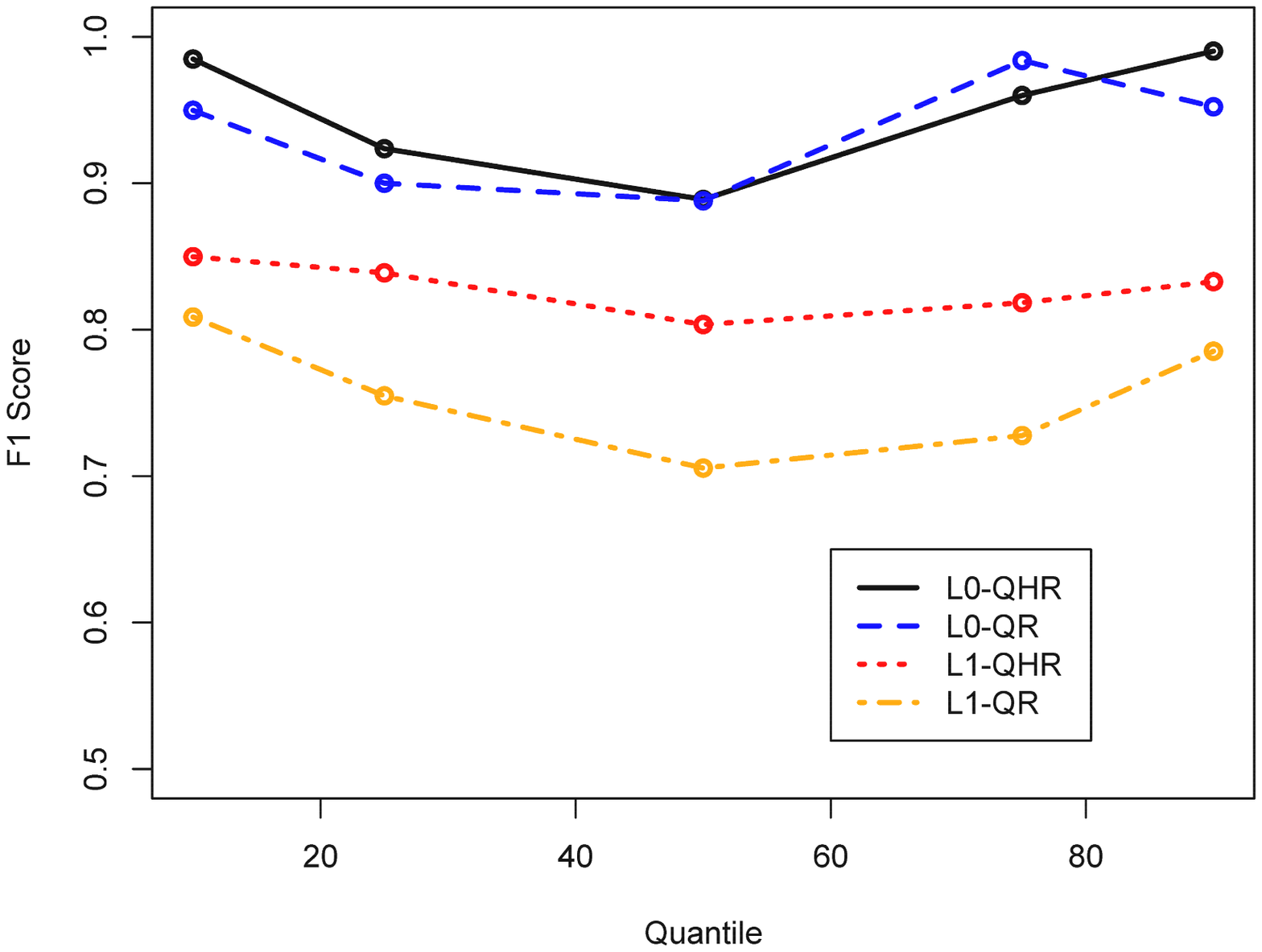}}\\
 $n=300,p=800$ 
\end{tabular}
  \caption{Variable Selection Accuracy ($F_1$ Score) for the comparison methods on simulated data as a function of the quantile. 
  $\ell_0$-QHR is represented using solid black, $\ell_0$-QR using dashed blue, $\ell_1$-QHR with dot red, and $\ell_1$-QR with dash-dot orange.
  Top: $n=300, p = 400$. Bottom: $n = 300, p = 800.$ 
 }
\label{fig:simulation1}
\end{figure}

We analyze quantile loss with $\tau$ selected from $\{0.1, 0.25, 0.5, 0.75, 0.9\}$, and for quantile Huber loss consider $\kappa\in\{.1, .2, 0.5, 1, 3\}$.
Figure~\ref{fig:simulation1} depicts the $F_1$ scores for the various methods as a function of the quantile considered.  From the figure we can make the following remarks:
\begin{itemize}
\item When the predictors do not necessarily impact the mean/median of the distribution it is critical to look at a wide spectrum of quantiles to capture all relevant predictors. 
\item Regardless of whether an OMP or $\ell_1$ regularized method is used, the quantile Huber loss yields higher accuracy than the quantile loss. 
This due to the fact that quantile Huber loss does not ``insist'' on fitting the \emph{inliers} exactly. 
\item Remarkably both $\ell_0$-QHR and $\ell_0$-QR achieve superior accuracy over $\ell_1$-QHR and $\ell_1$-QR.  
\end{itemize}

In comparison, the variable selection accuracy of Lasso is much lower, namely $F_1=0.56.$

\begin{figure}[h!]
\centering
\begin{tabular}{c}
\includegraphics[width=0.35\textwidth]{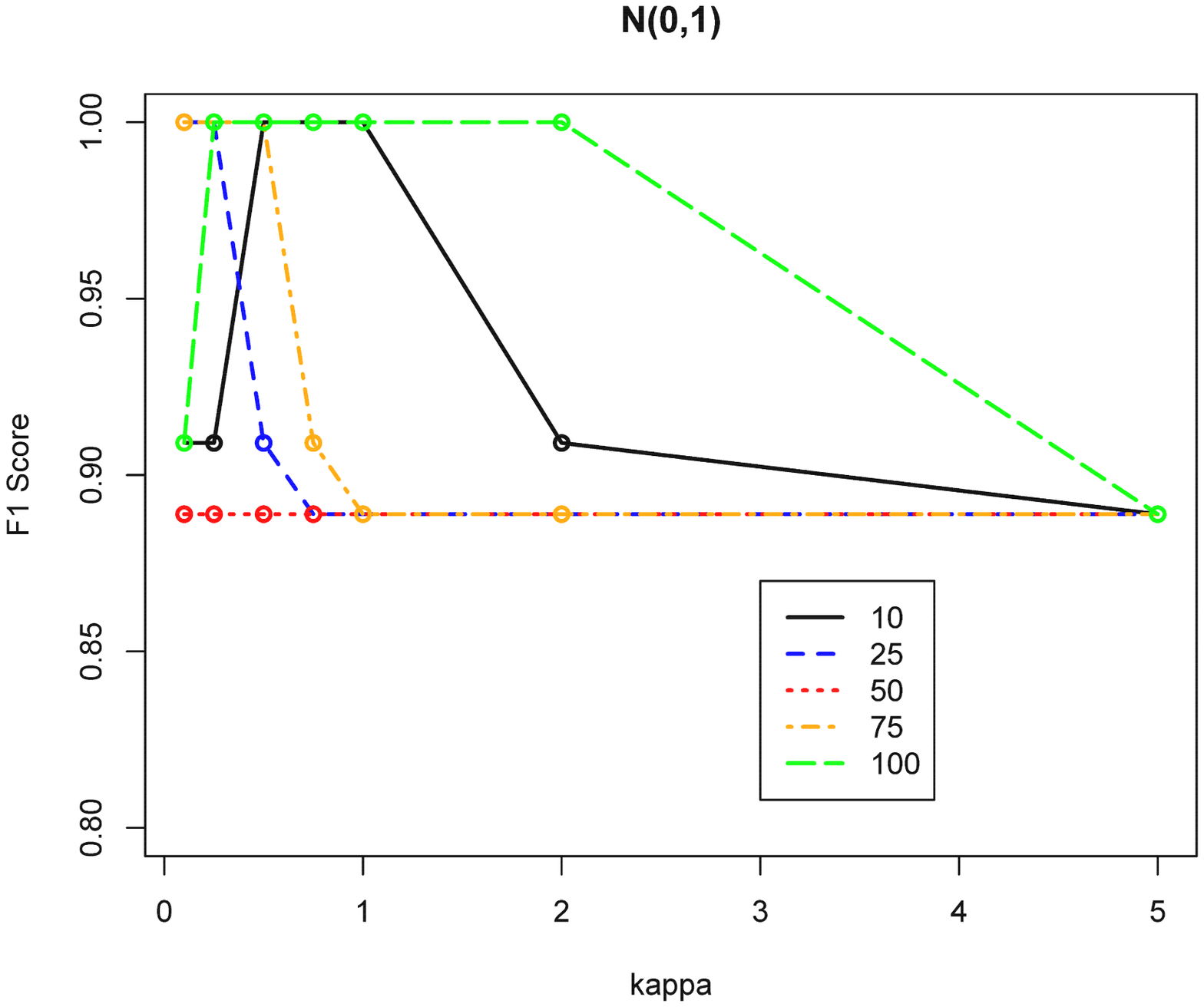} \\
\includegraphics[width=0.35\textwidth]{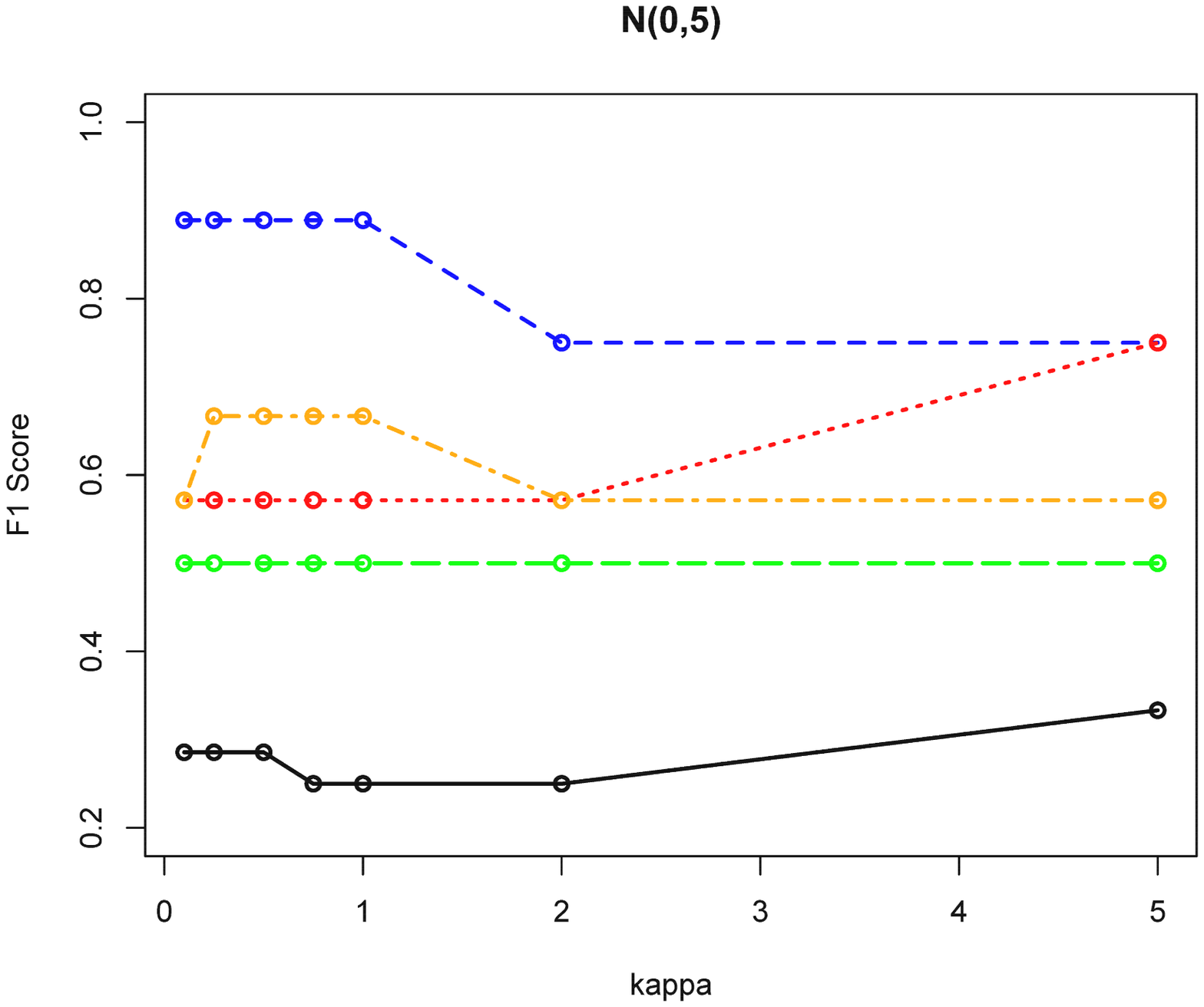}
\end{tabular}
  \caption{Variable Selection Accuracy ($F_1$ Score) for $\ell_0$-QHR on simulated data as a function of $\kappa$ for error $\sim N(0,1)$ (top) and error $\sim N(0,5)$ (bottom).}
 \label{fig:kappainfluence}
\end{figure}

We conclude the simulation study by briefly discussing the impact of $\kappa$ in the quantile Huber loss in terms of robustness. 
Figure~\ref{fig:kappainfluence} depicts the $F_1$ score for $\ell_0$-QHR and $\tau=0.25$ as a function of $\kappa$.  
Similar behavior is observed for $\ell_1$-QHR and is omitted due to space constraints.

\begin{figure*}[]
\centering
\begin{tabular}{ccc}
\includegraphics[width=55mm,height=45mm]{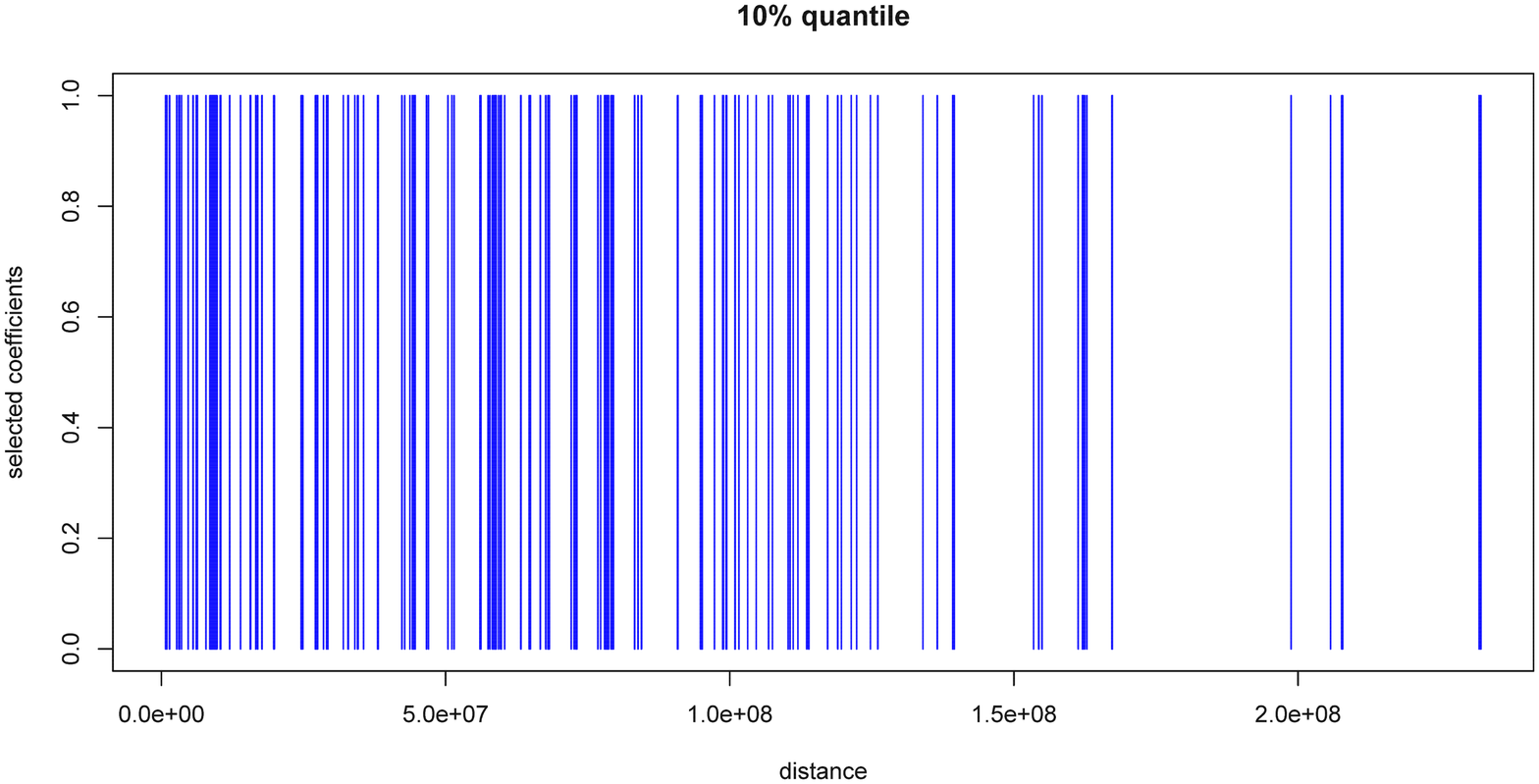} &
 \includegraphics[width=55mm,height=45mm]{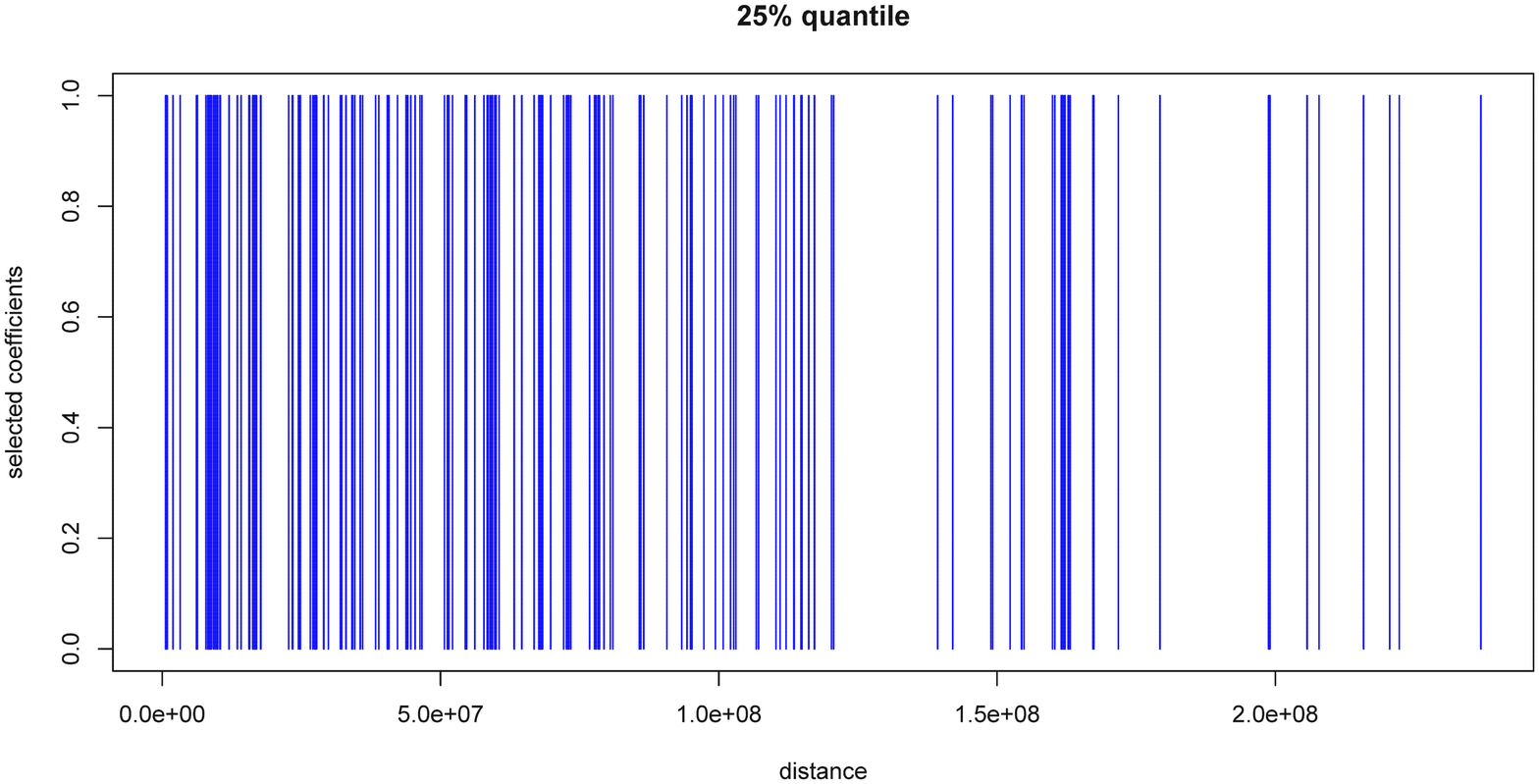}&
\includegraphics[width=55mm,height=45mm]{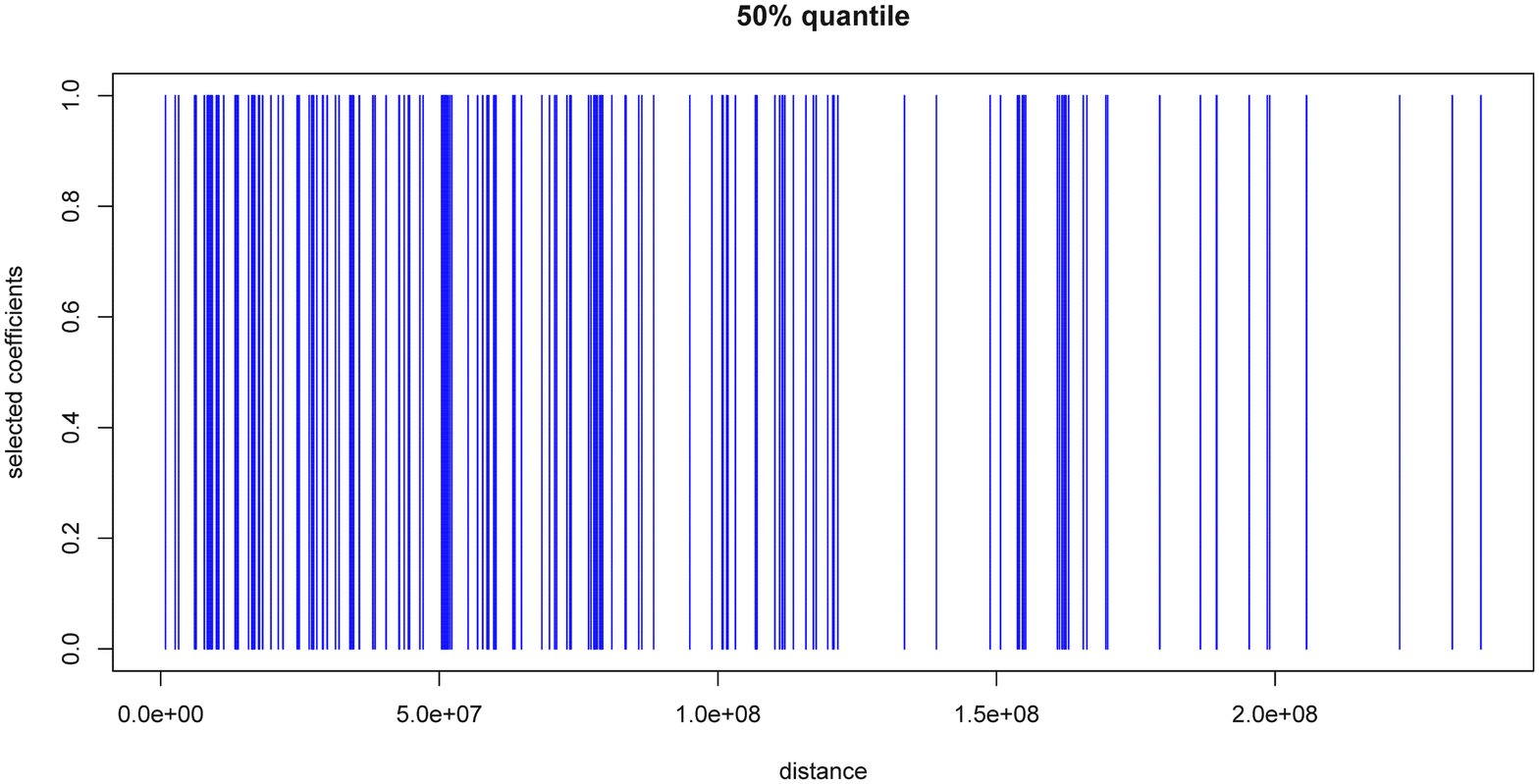} \\
 \includegraphics[width=55mm,height=45mm]{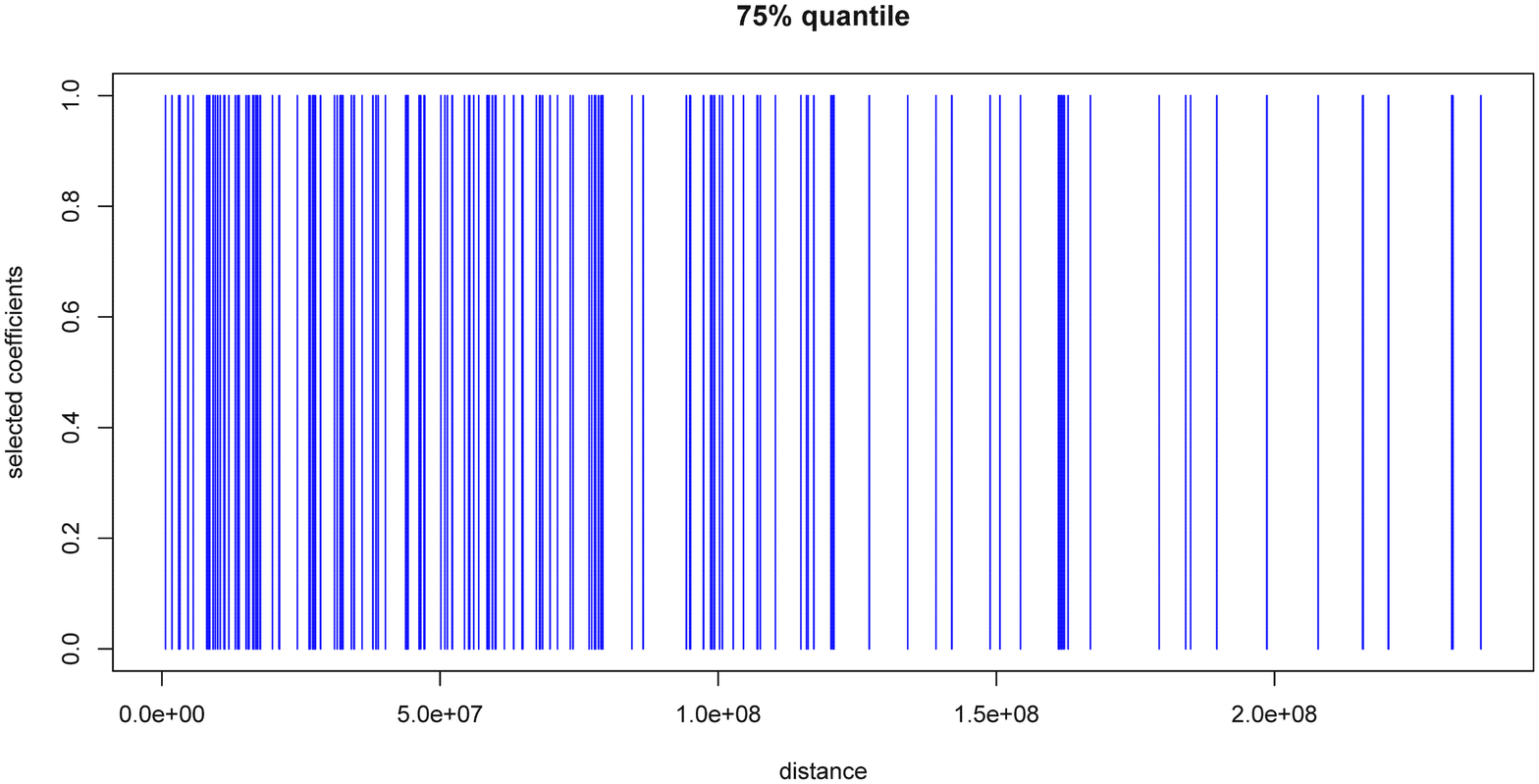}&
\includegraphics[width=55mm,height=45mm]{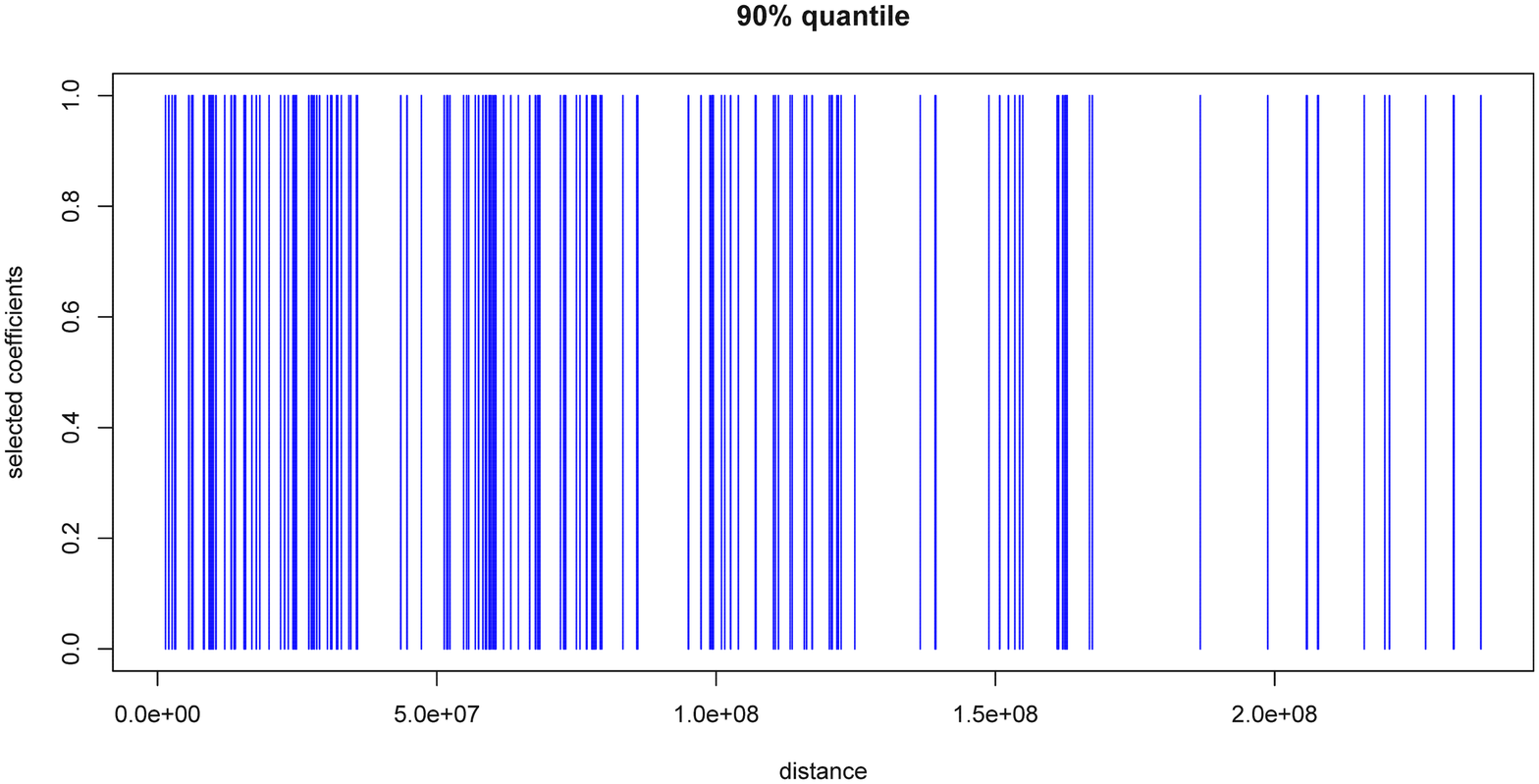} &  
\includegraphics[width=55mm,height=45mm]{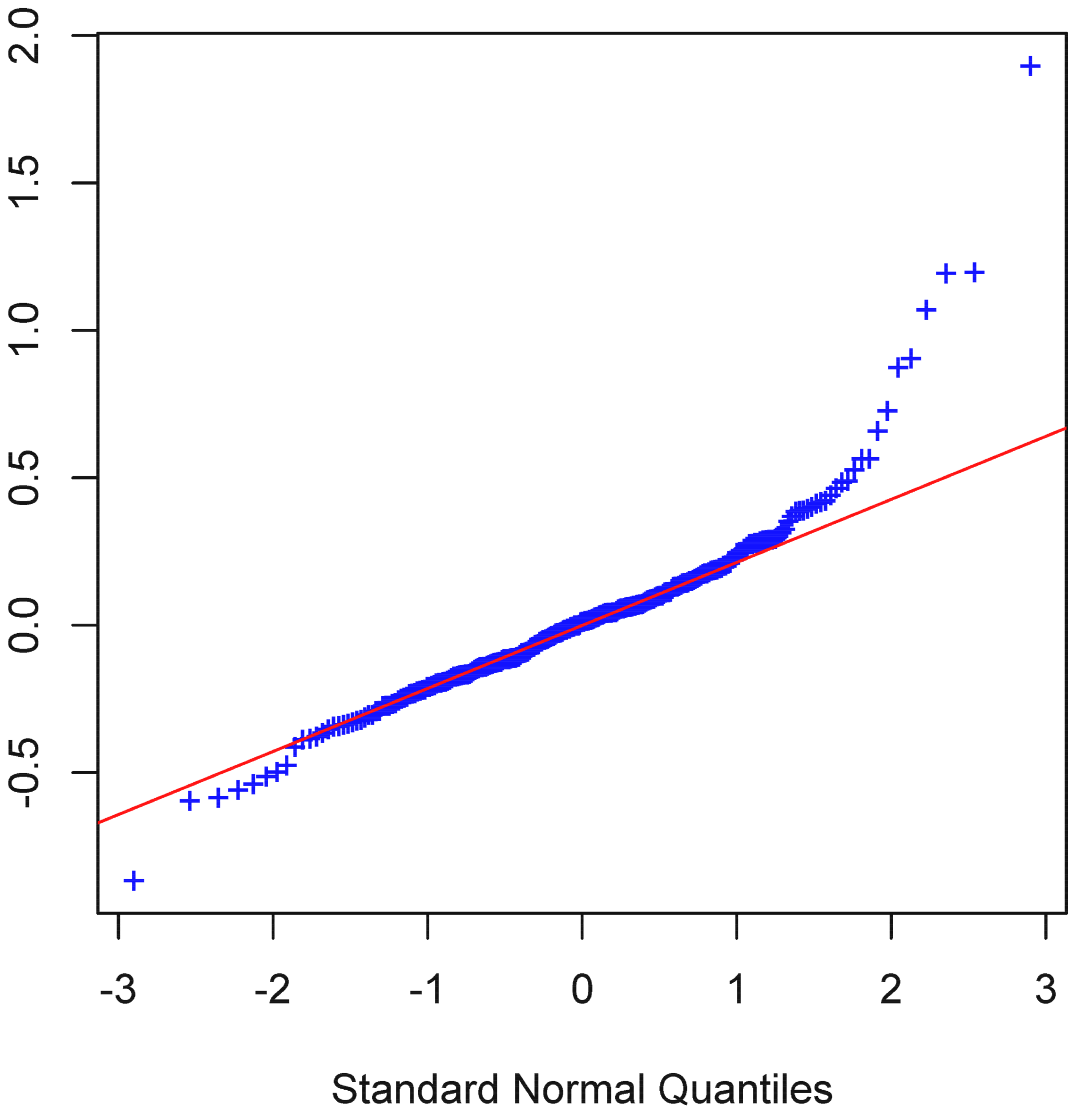}
\end{tabular}
  \caption{\label{fig:results_coef} 
Top $n$ regression coefficients of the SNPs, as estimated by $\ell_0$-QHR. The x-axis inidicates the distance (in kilobase) of the SNPs. From left to right, top to bottom: 10\% 25\%, 50\%, 75\%,90\% quantiles, and examplary QQ-plot of residuals}
\end{figure*}

Parameter $\kappa$ can be seen as a prior on the level of noise, providing a boundary between what is an acceptable level of noise and what is to be considered as outliers.  
Higher values of $\kappa$ are better suited for errors with lighter tails, while smaller values of $\kappa$ are more appropriate for heavier tails. 
A finer insight into the relationship between $\kappa$ and the error distribution is the subject of future work.
  
%

\subsection{Application to eQTL Mapping}

An interesting benefit of sparse quantile regression, in addition to its robustness to outliers, is the ability to perform variable selection in a quantile-dependent fashion. Sparse quantile regression can thus provide a more realistic picture of the sparsity patterns, which may be different at different quantiles.
We illustrate this point by analyzing Alzheimer's disease (AD) data generated by the Harvard Brain Tissue Resource Center and Merck Research Laboratories (http://sage.fhcrc.org/downloads/ downloads.php). This data concerns $n=206$ AD cases with SNPs and expression levels in the visual cortex. For our analysis, we selected $p=18137$ candidate SNPs based on a set of genes related to neurological diseases and studied the associations between these SNPs and the expression levels of the \emph{APOE} gene, which is a key Alzeimer's gene. Specifically, persons having an \emph{APOE e4} allele have an increased chance of developing the disease; those who inherit two copies of the allele are at even greater risk. 

Figure~\ref{fig:results_coef} shows the top $n$ SNPs  (sorted by the amplitude of their regression coefficient) selected by the $\ell_0$-QHR method for quantiles 10, 25, 50, 75, 90, as a function of the chromosome distance (similar insights can be gained from the $\ell_1$-QHR method, but we omit the plots due to space constraints.)

As can be seen from the figure, the coefficient profile indeed  changes based on the quantiles. This confirms the intuition that high-dimensional data such as those encountered in genomics is likely to exhibit heterogeneity due to some non-location-scale covariate effects. Sparse quantile regression enables a more comprehensive view of sparsity, where the set of relevant covariates can differ based on the segment of the conditional distribution under consideration. While some of the ``hotspots'' identified by our methods vary accross quantiles, it is also interesting to note that some do persist, in particular a hotspot within chromosome 19 where gene \emph{APOE} resides. 

To conclude the eQTL analysis, it is insightful to investigate the Normal QQ-plots of the residuals from the methods.  An examplary QQ-plot for the median quantile residuals of OMP is shown in Figure~\ref{fig:results_coef}.  We can see that the residuals have a very heavy right tail. This suggests that the robustness property of our methods is also valuable for this type of analysis.

\small
\bibliography{robust}

\begin{thebibliography}{40}
\providecommand{\natexlab}[1]{#1}
\providecommand{\url}[1]{\texttt{#1}}
\expandafter\ifx\csname urlstyle\endcsname\relax
  \providecommand{\doi}[1]{doi: #1}\else
  \providecommand{\doi}{doi: \begingroup \urlstyle{rm}\Url}\fi

\bibitem[Aravkin et~al.(2013)Aravkin, Burke, and
  Pillonetto]{AravkinBurkePillonetto2013}
Aravkin, Aleksandr~Y., Burke, James~V., and Pillonetto, Gianluigi.
\newblock Sparse/robust estimation and kalman smoothing with nonsmooth
  log-concave densities: Modeling, computation, and theory.
\newblock \emph{Journal of Machine Learning Research}, 14:\penalty0 2689--2728,
  2013.
\newblock URL \url{http://jmlr.org/papers/v14/aravkin13a.html}.

\bibitem[Bartlett \& Mendelson(2006)Bartlett and Mendelson]{bm-em-05}
Bartlett, Peter~L. and Mendelson, Shahar.
\newblock Empirical minimization.
\newblock \emph{Probability Theory and Related Fields}, 135\penalty0
  (3):\penalty0 311--334, 2006.

\bibitem[Beck \& Teboulle(2009)Beck and Teboulle]{Beck2009}
Beck, A. and Teboulle, M.
\newblock A fast iterative shrinkage-thresholding algorithm for linear inverse
  problems.
\newblock \emph{SIAM Journal of Imaging Sciences}, 2\penalty0 (1):\penalty0
  183--202, 2009.

\bibitem[Belloni \& Chernozhukov(2011)Belloni and Chernozhukov]{Belloni:2011}
Belloni, Alexandre and Chernozhukov, Victor.
\newblock L1-penalized quantile regression in high-dimensional sparse models.
\newblock \emph{Annals of Statistics}, 39\penalty0 (1):\penalty0 1012--1030,
  2011.

\bibitem[Buchinsky(1994)]{Buchinsky:1994}
Buchinsky, Moshe.
\newblock Changes in the u.s. wage structure 1963-1987: Application of quantile
  regression.
\newblock \emph{Econometrica}, 62\penalty0 (2):\penalty0 405--58, March 1994.

\bibitem[Cand{\`e}s \& Tao(2005)Cand{\`e}s and Tao]{CandesTao2005}
Cand{\`e}s, E.~J. and Tao, T.
\newblock Decoding by linear programming.
\newblock \emph{IEEE Transactions on Information Theory}, 51\penalty0
  (12):\penalty0 4203--4215, 2005.

\bibitem[Clark(1985)]{clark85}
Clark, D.
\newblock The mathematical structure of huber’s m-estimator.
\newblock \emph{SIAM Journal on Scientific and Statistical Computing},
  6:1:\penalty0 209--219, 1985.

\bibitem[Donoho(2006)]{Donoho2006_CS}
Donoho, D.
\newblock Compressed sensing.
\newblock \emph{IEEE Transactions on Information Theory}, 52\penalty0
  (4):\penalty0 1289--1306, 2006.

\bibitem[Fornasier(2006)]{Fornasier:06}
Fornasier, M.
\newblock Nonlinear projection recovery in digital inpainting for color image
  restoration.
\newblock \emph{J. Math. Imaging Vis.}, 24\penalty0 (3):\penalty0 359--373,
  2006.
\newblock ISSN 0924-9907.

\bibitem[Guyon \& Elisseeff(2003)Guyon and
  Elisseeff]{VariableSelection_JMLR_2003}
Guyon, I. and Elisseeff, A.
\newblock An introduction to variable and feature selection.
\newblock \emph{Journal of Machine Learning Research}, 3:\penalty0 1157--1182,
  2003.

\bibitem[J. et~al.(2004)J., Rosset, Hastie, and Tibshirani]{Zhu:2004}
J., Zhu., Rosset, S., Hastie, T., and Tibshirani, R.
\newblock 1-norm support vector machines.
\newblock \emph{Neural Information Processing Systems (NIPS)}, 16, 2004.

\bibitem[Johnson et~al.(2012)Johnson, Jalali, and Ravikumar]{Jalali12}
Johnson, Christopher~C., Jalali, Ali, and Ravikumar, Pradeep~D.
\newblock High-dimensional sparse inverse covariance estimation using greedy
  methods.
\newblock \emph{Proceedings of the 15th International Conference on Artificial
  Intelligence and Statistics}, 2012.

\bibitem[Kim et~al.(2007)Kim, Koh, Lustig, Boyd, and Gorinevsky]{Boyd2007}
Kim, Seung-Jean, Koh, K., Lustig, M., Boyd, S., and Gorinevsky, D.
\newblock An interior-point method for large-scale l1-regularized least
  squares.
\newblock \emph{Selected Topics in Signal Processing, IEEE Journal of},
  1\penalty0 (4):\penalty0 606--617, 2007.

\bibitem[Koenker(2005)]{Koenker:2005}
Koenker, R.
\newblock \emph{Quantile Regression}.
\newblock Cambridge University Press, 2005.

\bibitem[Koenker \& Bassett(1978)Koenker and Bassett]{KB78}
Koenker, R. and Bassett, G.
\newblock Regression quantiles.
\newblock \emph{Econometrica}, pp.\  33--50, 1978.

\bibitem[Koenker \& Geling(2001)Koenker and Geling]{KG01}
Koenker, R. and Geling, O.
\newblock Reappraising medfly longevity: A quantile regression survival
  analysis.
\newblock \emph{Journal of the American Statistical Association}, 96:\penalty0
  458–468, 2001.

\bibitem[Koenker \& Hallock(2001)Koenker and Hallock]{KH01}
Koenker, Roger and Hallock, Kevin~F.
\newblock Quantile regression.
\newblock \emph{Journal of Economic Perspectives, American Economic
  Association}, pp.\  143--156, 2001.

\bibitem[Li \& Swetits(1998)Li and Swetits]{LiW98}
Li, W. and Swetits, J.
\newblock The linear l1 estimator and the huber m-estimator.
\newblock \emph{SIAM Journal on Optimization}, 8\penalty0 (2):\penalty0
  457--475, 1998.

\bibitem[Li \& Zhu(2008)Li and Zhu]{Li2008}
Li, Y. and Zhu, J.
\newblock L1-norm quantile regression.
\newblock \emph{Journal of Computational and Graphical Statistics}, 17\penalty0
  (1):\penalty0 1--23, 2008.

\bibitem[Mallat \& Z(1993)Mallat and Z]{Mallat:1993}
Mallat, S.G and Z, Zhang.
\newblock Matching pusuits with time-frequency dictionaries.
\newblock \emph{IEEE Transcations on Signal Processing}, 41:\penalty0
  3397--3415, December 1993.

\bibitem[{Message Passing Interface Forum}(1995)]{MPI-1}
{Message Passing Interface Forum}.
\newblock {MPI}, June 1995.
\newblock \url{http://www.mpi-forum.org/}.

\bibitem[{Message Passing Interface Forum}(1997)]{MPI-2}
{Message Passing Interface Forum}.
\newblock {MPI}-2, July 1997.
\newblock \url{http://www.mpi-forum.org/}.

\bibitem[Neelamani et~al.(2010)Neelamani, Krohn, Krebs, Romberg, Deffenbaugh,
  and Anderson]{neelamani2010esf}
Neelamani, R., Krohn, C.~E., Krebs, J.~R., Romberg, J.~K., Deffenbaugh, Max,
  and Anderson, John~E.
\newblock Efficient seismic forward modeling using simultaneous random sources
  and sparsity.
\newblock \emph{Geophysics}, 75\penalty0 (6):\penalty0 WB15--WB27, 2010.

\bibitem[Ng(2004)]{Ng:2004}
Ng, A.
\newblock Feature selection, l1 vs. l2 regularization, and rotational
  invariance.
\newblock \emph{Proceedings of 21st International Conference on Machine
  Learning (ICML)}, 2004.

\bibitem[Portnoy \& Koenker(1997)Portnoy and Koenker]{Koenker:1997}
Portnoy, Stephen and Koenker, Roger.
\newblock The gaussian hare and the laplacian tortoise: Computability of
  squared- error versus absolute-error estimators.
\newblock \emph{Statistical Science}, 12\penalty0 (4):\penalty0 pp. 279--296,
  1997.

\bibitem[Poulson et~al.(2013)Poulson, Marker, van~de Geijn, Hammond, and
  Romero]{Poulson:2012:ENF}
Poulson, Jack, Marker, Bryan, van~de Geijn, Robert~A., Hammond, Jeff~R., and
  Romero, Nichols~A.
\newblock Elemental: A new framework for distributed memory dense matrix
  computations.
\newblock \emph{{ACM} Transactions on Mathematical Software}, 39\penalty0
  (2):\penalty0 13:1--13:24, February 2013.
\newblock URL \url{http://doi.acm.org/10.1145/2427023.2427030}.

\bibitem[Rockafellar \& Wets(1998)Rockafellar and Wets]{RTRW}
Rockafellar, R.T. and Wets, R.J.B.
\newblock \emph{Variational Analysis}, volume 317.
\newblock Springer, 1998.

\bibitem[Rosset \& Zhu(2007)Rosset and Zhu]{Ross2007}
Rosset, S. and Zhu, J.
\newblock Piecewise linear regularized solution paths.
\newblock \emph{Annals of Statistics}, 35\penalty0 (3):\penalty0 1012--1030,
  2007.

\bibitem[Shalev-Shwartz et~al.(2010)Shalev-Shwartz, Srebro, and Zhang]{siopt10}
Shalev-Shwartz, Shai, Srebro, Nathan, and Zhang, Tong.
\newblock Trading accuracy for sparsity in optimization problems with sparsity
  constraints.
\newblock \emph{Siam Journal on Optimization}, 20:\penalty0 2807--2832, 2010.

\bibitem[Starck et~al.(2005)Starck, Elad, and Donoho]{MCA3-2005}
Starck, J.-L, Elad, M., and Donoho, D.
\newblock Image decomposition via the combination of sparse representation and
  a variational approach.
\newblock \emph{IEEE Transaction on Image Processing}, 14\penalty0 (10), 2005.

\bibitem[Tibshirani(1996)]{Lasso1996}
Tibshirani, R.
\newblock Regression shrinkage and selection via the {LASSO}.
\newblock \emph{Journal of the Royal Statistical Society, Series B.},
  58\penalty0 (1):\penalty0 267--288, 1996.

\bibitem[Wakin et~al.(2006)Wakin, Laska, Duarte, Baron, Sarvotham, Takhar,
  Kelly, and Baraniuk]{wakin2006civ}
Wakin, M., Laska, J., Duarte, M., Baron, D., Sarvotham, S., Takhar, D., Kelly,
  K., and Baraniuk, R.
\newblock {Compressive imaging for video representation and coding}.
\newblock \emph{Proc. Picture Coding Symposium}, 2006.

\bibitem[Wang et~al.(2007)Wang, Li, and Tsai]{AR_LASSO_2007}
Wang, H., Li, G., and Tsai, C.L.
\newblock Regression coefficient and autoregressive order shrinkage and
  selection via the lasso.
\newblock \emph{Journal Of The Royal Statistical Society Series B}, 69\penalty0
  (1):\penalty0 63--78, 2007.

\bibitem[Wang et~al.(2012)Wang, Wu, and Li]{Wang:2012}
Wang, Lan, Wu, Yichao, and Li, Runze.
\newblock Quantile regression for analyzing heterogeneity in ultra-high
  dimension.
\newblock \emph{Journal of the American Statistical Association}, 107\penalty0
  (497):\penalty0 214--222, 2012.

\bibitem[Wright(1997)]{Wright:1997}
Wright, S.J.
\newblock \emph{Primal-dual interior-point methods}.
\newblock Siam, Englewood Cliffs, N.J., USA, 1997.

\bibitem[Ye \& Anstreicher(1993)Ye and Anstreicher]{Ye:1993}
Ye, Yinyu and Anstreicher, Kurt.
\newblock On quadratic and $o(\sqrt{nL})$ convergence of a predictor-corrector
  method for lcp.
\newblock \emph{Mathematical Programming}, 62\penalty0 (1-3):\penalty0
  537--551, 1993.

\bibitem[Zhang(2008)]{Zhang:2008}
Zhang, T.
\newblock Adaptive forward-backward greedy algorithm for sparse learning with
  linear models.
\newblock \emph{Neural Information Processing Systems (NIPS)}, 21, 2008.

\bibitem[Zheng(2011)]{Zheng2011}
Zheng, Songfeng.
\newblock Gradient descent algorithms for quantile regression with smooth
  approximation.
\newblock \emph{International Journal of Machine Learning and Cybernetics},
  2\penalty0 (3):\penalty0 191--207, 2011.

\bibitem[Zou(2006)]{Zou2006}
Zou, H.
\newblock The adaptive lasso and its oracle properties.
\newblock \emph{Journal of the American Statistical Association}, 101:\penalty0
  1418--1429, 2006.

\bibitem[Zou \& Yuan(2008)Zou and Yuan]{Zou08}
Zou, Hui and Yuan, Ming.
\newblock Regularized simultaneous model selection in multiple quantiles
  regression.
\newblock \emph{Computational Statistics {\&} Data Analysis}, 52\penalty0
  (12):\penalty0 5296--5304, 2008.

\end{thebibliography}
\bibliographystyle{icml2014}

\end{document}


\maketitle

\section{Proof of Theorem 1}
\textit{Proof:} Let $\bar x$ be the true model parameter, and $\Delta=x-\bar x.$ The first step of the proof consists in approximating the $\ell_1$ penalized quantile Huber loss $L_n(x) + \lambda_n \|x\|_1$ around $\bar x$ using {\small
\[
M_n( x)= \sum_{i=1}^n  (1\{|\epsilon_i|\geq \kappa\} \kappa f(0) + 1\{|\epsilon_i|< \kappa\}) (A_i'\Delta)^2 - \sum_{i=1}^n  (1\{|\epsilon_i|\geq \kappa\}\kappa \eta_i + 1\{|\epsilon_i|< \kappa\}\epsilon_i) A_i \Delta + \lambda_n \| x\|_1,
\] }where $\eta_i=\textrm{sign}(\epsilon_i).$

Let $R_{i,n}(x)=\rho_{0.5}(\epsilon_i-\Delta'A_i)- \rho_{0.5}(\epsilon_i) + (1\{|\epsilon_i|\geq \kappa\} \kappa\eta_i + 1\{|\epsilon_i|< \kappa\}\epsilon_i) A_i \Delta.$
There holds 
\[ 
|L_n(x)+ \lambda_n \|x\|_1 - L_n(\bar x) - M_n(x)| = o(\|A \Delta\|_2^2) + \sum_{i=1}^n \xi_{in}(x).\]
where $\xi_{i,n}(x)=R_{i,n}(x)-E_{\epsilon}R_{i,n}(x).$

Let $\tilde x_n=\arg\min _x M_n( x).$ We note that $\tilde x_n$ is the Least-squares Lasso estimator of a modified regression model with modified predictors $\tilde A_{ij}=\sqrt{(1\{|\epsilon_i|\geq \kappa\}\kappa f(0) + 1\{|\epsilon_i|< \kappa\})} A_{ij}$ and modified error terms $\tilde \epsilon_i= (2     \sqrt{(1\{|\epsilon_i|\geq \kappa\} \kappa f(0) + 1\{|\epsilon_i|< \kappa\})})^{-1}(1\{|\epsilon_i|\geq \kappa\} \kappa \eta_i + 1\{|\epsilon_i|< \kappa\}\epsilon_i).$
For the same $\lambda_n$ denote by $B$ the union of supports of $\tilde x(\lambda_n),$ $\hat x(\lambda_n)$ and $\bar x.$
For any $\delta>0$ let $S_\delta^o= \{ x_B \in R^{|B|} : \| x_B - \tilde  x_B\|_2=\delta\}$ and $S_\delta= \{ x_B \in R^{|B|} : \| x_B - \tilde  x_B\|_2\leq\delta.\}$ Define $h_n(\delta)=\inf_{ x_B \in S_{\delta}^o} n^{-1} M_n ( x_B) - n^{-1} M_n (\tilde x_B)$ and $\Delta_n(\delta)=\sup_{ x_B \in S_\delta} n^{-1} |L_n( x_B) + \lambda_n \| x_B\|_1 - L_n(\bar x_B) - M_n( x_B)|.$
Since $\| x_B-\bar x_B\|_2^2 \leq 2 \| x_B - \tilde x_B\|_2^2 + 2 \|\tilde x_B- \bar x\|_2^2$ we have:
$\Delta_n(\delta) \leq o_p(\delta^2 \kappa_u) +o_p(\sup_{ x_B \in S_\delta}\|\tilde x_B-\bar x_B\|_2^2) + \sup_{ x_B \in S_\delta} | \sum_{i=1}^n \xi_{i,n}( x_B)|,$ and $h_n(\delta) \geq \inf_{ x_B \in S_\delta^o} \frac{1}{n} \sum_{i=1}^n (1\{|\epsilon_i|\geq \kappa\}\kappa f(0) + 1\{|\epsilon_i|< \kappa\}) (A_{i,B}'( x_B-\tilde x_B))^2 \geq \delta^2 \kappa_l \min(1,\kappa f(0)) >0.$ Let $\delta_1=\delta/2>0.$ From the convex minimization theorem in \cite{HjortPollard93} we get
\begin{eqnarray}
P(\|\hat x_B - \tilde x_B\|_2 > \delta_1) 
& \leq& P(\Delta_n(\delta_1) > h_n(\delta_1)/2)\nonumber\\
&\leq & P(\sup_{ x_B \in S_\delta^d} |n^{-1} \sum_{i=1}^n \xi_{i,n}( x_B)| > \frac{\delta_1^2 \min(1,\kappa f(0)) \kappa_l}{4}) \nonumber \\&&+ P (\|\tilde x_B - \bar x_B\|_2>\delta_1)\nonumber\\
&=&P(Z(\delta)> \frac{\delta_1^2 \min(1,\kappa f(0)) \kappa_l}{4}) + P (\|\tilde x_B - \bar x_B\|_2>\delta_1),\label{eq:al:bound1}
\end{eqnarray}
where $Z(\delta)=\sup_{ x_B \in S_\delta^d} |n^{-1} \sum_{i=1}^n \xi_{i,n}( x_B)| .$

Let $f_{ x_B}(A_{i,B})=( x_B - \bar x_B)'A_{i,B}.$ Clearly $f_{\bar x_B}(A_{i,B})=0.$ Let $v_i(\epsilon_i,t)=\int_0^t [I(0<\epsilon_i \leq \nu )] d\nu.$ There holds $|v_i(\epsilon_i,t_2)-v_i(\epsilon_i,t_1)| \leq |t_2-t_1|.$

If $|\epsilon_i|>\kappa$ then $R_{i,n}( x_B)=2\kappa v_i(f_{ x_B}(A_{i,B},\epsilon_i).$ On the other hand if $|\epsilon_i| \leq \kappa$ then $R_{i,n}( x_B)=\frac{1}{2} (A_{i,B} '\Delta_B)^2.$  Let $\mu_i$'s be a Rademacher sequence.

We have \begin{eqnarray*}
E_\epsilon Z(\delta) &=& E_\epsilon \sup_{ x_B \in S_\delta^d} n^{-1} |\sum_{i=1}^n [R_{i,n}( x_B)-E_\epsilon(R_{i,n}( x_B))|\\
&\leq& 2 E_\epsilon \sup_{ x_B \in S_\delta^d} n^{-1}\left|\sum_{i=1}^n \mu_i \left(1[|\epsilon_i| \geq \kappa] (2\kappa v_i(f_{ x_B}(A_{i,B}))-2\kappa v_i(f_{\bar x_B}(A_{i,B})))  \right.\right. \\&& \left. \left.+ 1[|\epsilon_i| \leq \kappa] (0.5 f_{ x_B}(A_{i,B}))^2\right)\right|\\
&\leq& 8 \kappa E_\epsilon \sup_{ x_B \in S_\delta^d} n^{-1} \left| \sum_{i=1}^n 1[|\epsilon_i| \geq \kappa] \mu_i (f_{ x_B}(A_{i,B})-f_{\bar x_B}(A_{i,B}))\right| \\&&+ 2 M \sqrt {|B|} \delta E_\epsilon \sup_{ x_B \in S_\delta^d} n^{-1} \left| \sum_{i=1}^n 1[|\epsilon_i| < \kappa] \mu_i (f_{x_B}(A_{i,B})-f_{\bar x_B}(A_{i,B}))\right|\\
&\leq& (8 \sqrt{|B|} \delta \kappa \pi_\kappa + 2 |B| M\delta^2 (1-\pi_\kappa))\left(\sqrt{2n^{-1} \log(2|B|)}+ M n^{-1} \log(2|B|)\right).
\end{eqnarray*}

Let $\iota_n=\max(4M\kappa \sqrt{|B|} \delta,M^2 |B| \delta^2).$ Let $\kappa_n^2=4c^u \delta^2(\pi_\kappa \kappa +(1-\pi_\kappa)\frac{M^2}{4}|B|\delta^2).$
Then by the concentration theorem of \cite{BousquetKP02} we have for any $z >0$ \[P(Z(\delta)\geq EZ(\delta) + z \sqrt{2(\kappa_n^2 + \iota_n EZ(\delta))} + \frac{z^2 \iota_n}{3}) \leq \exp(-nz^2).\]

Set $\delta=\alpha d_{n} \frac{\log p}{\kappa_l^2 n} \kappa^2 \left(\frac{\pi_\kappa }{2\kappa f(0)} + 2(1-\pi_\kappa)\right),$ for some constant $\alpha>0.$ By choosing $z$ appropriately, and combining with \eqref{eq:al:bound1} we get 
\begin{equation}\lim_{n\to\infty} P(\|\hat x_B - \tilde x_B\|_2 > \delta_1)  \leq \lim_{n\to\infty} P (\|\tilde x_B - \bar x_B\|_2>\delta_1).\label{eq:al:bound2}\end{equation}

Let $\tilde B$ denote the support of $\tilde x.$ From the study of the traditional Least-Squares Lasso in \cite{Meinshausen09} we get $\tilde B \leq \kappa_u \left(\frac{n}{\lambda}\right)^2 \kappa^2\left(\frac{\pi_\kappa}{2\kappa f(0)} + 2(1-\pi_\kappa)\right).$
Similarly let $\hat B$ denote the support of $\hat  x.$ 
By examining the KKT conditions of our objective function we  obtain that $|\hat B| \leq 2 \kappa^2 \kappa_u \left(\frac{n}{\lambda}\right)^2.$
Thus $|B| \leq (c_1 + ... ) \left(\frac{n}{\lambda}\right)^2.$
Thus $| \tilde B| + s  \leq \left(c_1 + \kappa_u\kappa^2\left(\frac{\pi_\kappa}{2\kappa f(0)} + 2(1-\pi_\kappa)\right)\right) \left(\frac{n}{\lambda}\right)^2.$

By adapting the reasoning in \cite{Meinshausen09}  for the traditional lasso we obtain {\small
\[\| \tilde x_B - \bar x_B\|_2^2 \leq O\left(\left(\frac{\lambda_n}{\kappa_l n}\right)^2 s \kappa^2 \left(\frac{\pi_\kappa }{2\kappa f(0)} + 2(1-\pi_\kappa)\right)\right) +O_P\left(d_{n} \frac{\log p}{\kappa_l^2 n} \kappa^2 \left(\frac{\pi_\kappa }{2\kappa f(0)} + 2(1-\pi_\kappa)\right)\right),
\]}
where $d_{n}=\left(c_1 + \kappa_u\kappa^2\left(\frac{\pi_\kappa}{2\kappa f(0)} + 2(1-\pi_\kappa)\right)\right) \left(\frac{n}{\lambda_n}\right)^2.$ Combining the above with~\eqref{eq:al:bound2}  yields the theorem statement. \hfill $\Box$

\bibliography{robust}
\bibliographystyle{plain}